%% file: main_cpinn.tex
\documentclass{article}

\PassOptionsToPackage{square,numbers,sort&compress}{natbib}
\def\biblio{\bibliographystyle{abbrvnat}\bibliography{refs}}
\usepackage[preprint]{neurips_2022}

\usepackage[utf8]{inputenc} 
\usepackage[T1]{fontenc}    
\usepackage{hyperref}       
\usepackage{url}            
\usepackage{booktabs}       
\usepackage{amsfonts}       
\usepackage{nicefrac}       
\usepackage{microtype}      
\usepackage{xcolor}         
\usepackage[pdftex]{graphicx}
\usepackage{enumitem}
\graphicspath{{images/}}

\usepackage{amssymb}
\usepackage{amsmath}
\usepackage{latexsym}
\usepackage{mathtools}
\usepackage{mleftright}

\usepackage{url}
\definecolor{newcolor}{rgb}{.8,.349,.1}
\usepackage{xr-hyper}
\usepackage{subfiles}
\usepackage{hyperref}
\usepackage[ruled,vlined]{algorithm2e}
\usepackage{graphicx,pgfplots,tikz}
\usepackage{listings}
\usepackage{xcolor}
\graphicspath{{./images/}}
\usepackage{float}
\usepackage{physics}
\usepackage{cleveref}
\usepackage[font=bf]{caption}
\usepackage{wrapfig}
\usepackage{algorithmic}
\usepackage{listofitems} 
\usepackage{todonotes}
\usepackage{subcaption}
\usepackage{tabularx}
\usepackage{soul}
\usepackage{csl}
\usepackage{wrapfig}
\usepackage{booktabs}

\usepackage[listings,skins,breakable]{tcolorbox}

\DeclareMathOperator*{\argmin}{arg\,min}

\hypersetup{
    colorlinks=true,
    linkcolor=blue,
    filecolor=magenta,      
    urlcolor=cyan,
}

\def\!#1{\mathcal{#1}}
\def\*#1{\boldsymbol{\mathbf{#1}}}
\def\|#1{\textnormal{#1}}
\def\##1{\mathfrak{#1}}

{%
\end{tcolorbox}\endgroup}

\title{Physics-informed neural networks for PDE-constrained optimization and control}

\author{%
  Jostein Barry-Straume\thanks{\url{https://github.com/ComputationalScienceLaboratory/control-pinns}} \\
  Computational Science Laboratory \\
  Department of Computer Science \\
  Blacksburg, VA 24061 \\
  \texttt{jostein@vt.edu} \\
  \And
  Arash Sarshar \\
  Computational Science Laboratory \\
  Department of Computer Science \\
  Blacksburg, VA 24061 \\
  \texttt{sarshar@vt.edu} \\
  \And
  Andrey A. Popov \\
  Computational Science Laboratory \\
  Department of Computer Science \\
  Blacksburg, VA 24061 \\
  \texttt{apopov@vt.edu} \\
  \And
  Adrian Sandu\thanks{\url{https://csl.cs.vt.edu/}} \\
  Computational Science Laboratory \\
  Department of Computer Science \\
  Blacksburg, VA 24061 \\
  \texttt{sandu@cs.vt.edu} \\
}

\begin{document}
\def\biblio{}

\csltitle{Physics-informed neural networks for PDE-constrained optimization and control}

\cslauthor{Jostein Barry-Straume, Arash Sarshar,\\Andrey A. Popov, and Adrian Sandu}

\cslyear{22}
\cslreportnumber{2}
\csltitlepage


\begin{abstract}


A fundamental problem in science and engineering is designing optimal control policies that steer a given system towards a desired outcome.
This work proposes Control Physics-Informed Neural Networks (Control PINNs) that simultaneously solve for a given system state, and for the optimal control signal, in a one-stage framework that conforms to the underlying physical laws. 
Prior approaches use a two-stage framework that first models and then controls a system in sequential order. In contrast, a Control PINN incorporates the required optimality conditions in its architecture and in its loss function. 
The success of Control PINNs is demonstrated by solving the following open-loop optimal control problems: (i) an analytical problem, (ii) a one-dimensional heat equation, and (iii) a two-dimensional predator-prey problem.
\end{abstract}

\subfile{sections/introduction}

\subfile{sections/appendix_open_loop_approach}

\subfile{sections/appendix_survey}

\subfile{sections/methodology}

\subfile{sections/analytical_problem}

\subfile{sections/heat_equation}

\subfile{sections/predator_prey}

\subfile{sections/conclusion}

\begin{ack}
Special thanks to Austin Chennault for pointing out many relevant literature to this project. This work was supported by DOE through award ASCR DE-SC0021313, by NSF  through award CDS\&E--MSS 1953113, and by the Computational Science Laboratory at Virginia Tech. All data and codes used in this manuscript are publicly available on GitHub at: \url{https://github.com/ComputationalScienceLaboratory/control-pinns}.
\end{ack}

\newpage
\bibliography{refs}
\newpage
\appendix

\subfile{sections/appendix_analytical_problem}

\subfile{sections/appendix_heat_equation}

\subfile{sections/appendix_predator_prey}

\end{document}

%% file: sections/introduction.tex
\section{Introduction}
\label{sec:Introduction}

Scientific Machine Learning (SciML) has arisen as a replacement to traditional numerical discretization methods. The main driving force behind this replacement is neural networks (NN), largely due to their success in natural language processing  and computer vision  problems \cite{atzberger2018importance}. 
As a vehicle to approximate the solution to a given partial differential equation (PDE) or ordinary differential equation (ODE), NNs offer a mesh-free approach via auto differentiation, and break the curse of dimensionality \cite{lu2021deepxde}.
Combining scientific computing and ML, SciML offers the potential to improve ``predictions beyond state-of-the-art physical models with smaller number of samples and generalizability in out-of-sample scenarios'' \cite{willard2021integrating}.     
Willard \textit{et} al. provide a structured overview of physics-based modeling approaches with ML techniques, and summarize current areas of application with regard to science-guided ML in \cite{willard2021integrating}.

Physics-Informed Neural Networks (PINNs) \cite{raissi2019physics} solve semi-supervised learning tasks while respecting the properties of physical laws. This is achieved by informing the loss function about the mathematical equations that govern the physical system. Raissi \textit{et} al. utilize PINNs for solving physical equations, and for data-driven discovery of partial differential equations \cite{raissi2019physics}. The general procedure for solving a differential equation with a PINN involves finding the parameters of a network that minimize a loss function involving the mismatch of output and data, as well as residuals of the boundary and initial conditions, PDE equations, and any other physical constraints required \cite{lu2021deepxde}. The recent survey paper by Cuomo \textit{et} al. \cite{cuomo2022scientific} provide a comprehensive overview of PINNS and discusses a variety of customizations  ``through different activation functions, gradient optimization techniques, neural network structures, and loss functions''.
Since their introduction, PINNs have been leveraged to solve a wide range of problems including, but not limited to, solution of PDEs \cite{raissi2019physics, Kollmannsberger2021}, inverse problems \cite{lu2021deepxde, chen2020, QiZhi2020, raissi2017physicsI, raissi2017physicsII, raissi2020}, solution of fractional differential equations \cite{pang2019}, and stochastic differential equations \cite{zhang2019, yang2018physicsinformed, nabian2019, zhang2019learning}. 


\begin{wrapfigure}{l}{0.45\textwidth}
    \centering
    \includegraphics[width=0.44\textwidth]{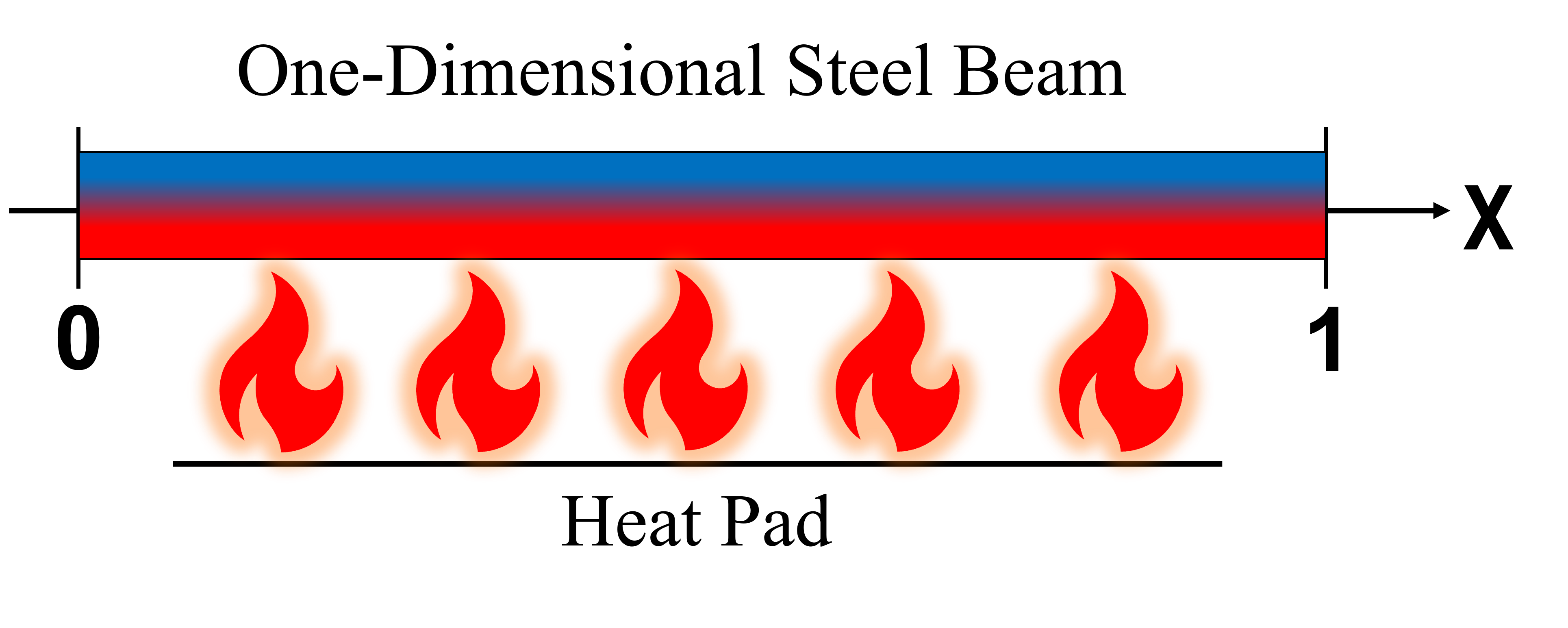}
    \caption{\textmd{\emph{Explanatory diagram:} A one-dimensional steel beam is heated by a controllable heat pad over the space $x$ from $x_0$ to $x_f$ such that at the final time a given temperature of the beam will be realized.}}
    \label{fig:steel-beam}
\end{wrapfigure}

Optimal control problems are multistage decision processes in which numerous decisions over time and space are made sequentially for a given system \cite{bellman1966dynamicprog}. 
This sequential repetition of decisions can be thought of as a control policy that, given an initial state of the system, efficiently enacts all remaining decisions such that an objective is either minimized or maximized (e.g. minimizing time, or maximizing profit) \cite{bellman1966dynamicprog}. 
Knowledge of both the system state and the effects of the control are necessary to develop an optimal control policy via learning through experience. 
Some recent work has began exploring the application of PINNs to solve optimal control problems \cite{hwang2021solving, chen2019optimal, antonelo2021physicsinformed, mowlavi2021optimal, sifan2021deeponet}.
However, these existing approaches either feed control data to a PINN, thereby training on precomputed control signals, or use an external control mechanism in conjunction with the PINN model of the physical system.

This paper presents a new PINN framework, named Control PINNs, for solving open-loop optimal control problems. Control PINNs simultaneously solve the learning tasks of the system state, the adjoint system state, and the optimal control signal, without the need for either a priori controller data or an external controller. In other words, Control PINNs can find optimal control solutions to complex computational scientific problems without the need for dividing the learning tasks of the system state and the control signal into separate learning-through-experience frameworks. \Cref{sec:Appendix-Open-Loop-Approach} details the advantages and disadvantages between the sequential and simultaneous learning approaches. A one-stage framework, such as Control PINN, is desirable because they solve PDEs in continuous time and space, and avoid the huge systems of nonlinear equations created by two-stage frameworks \cite{biegler2003large}.

The approaches in \cite{hwang2021solving,chen2019optimal,antonelo2021physicsinformed, mowlavi2021optimal, sifan2021deeponet} build accurate PINNs and use them as differentiable surrogate models in the optimization solution for control applications. For a full survey of \cite{hwang2021solving,chen2019optimal,antonelo2021physicsinformed, mowlavi2021optimal, sifan2021deeponet}, please refer to \cref{sec:Appendix-Survey}. 
Going beyond these prior approaches, we propose a framework that can be considered a new taxonomical entity within the genus of PINNs, wherein the PDE-constrained optimization is fused with the training loss function to create a one-stage approach to directly learning both the system solution and the optimal control.

The remainder of the paper is organized as follows. 
\Cref{sec:Appendix-Open-Loop-Approach} delves into the advantages and disadvantages of one-stage versus two-stage frameworks.
\Cref{sec:Appendix-Survey} offers a survey of current approaches in optimal control with PINNs.
\Cref{sec:Methodology} covers the methodology of the novel approach that this paper offers. \Cref{sec:analytical-toy-problem} validates the methodology via implementation of an analytical problem. \Cref{sec:1d-heat-equation} presents and discusses experimental results of a one-dimensional heat equation. \Cref{sec:2d-predator-prey} offers a more challenging optimal control problem  for  a two-dimensional predator-prey reaction-diffusion problem. \Cref{sec:Conclusion} summarizes the results and details the groundwork for future directions.

\biblio

%% file: sections/appendix_open_loop_approach.tex
\section{Open-loop control problems}
\label{sec:Appendix-Open-Loop-Approach}

Two general approaches exist to solve open-loop optimal control problems \cite{biegler2003large}. The first is nested analysis and design (NAND), which is also known as black-box or two-stage approach. The heuristic for this approach is: (i) start with a certain initial guess of the optimal control $\*u$ and optimal solution $\*y$, (ii) solve the forward equations and evaluate the loss function, (iii) solve the adjoint equations to evaluate the gradient with respect to controls, (iv) update the controls using an optimization scheme such as BFGS, and (v) repeat the process for the new controls.

The advantage of this approach is the relative ease of implementation. The disadvantages are the requirement to resolve the model equations at each iteration of the optimization algorithm in order to determine the state and adjoint, the fact that the optimizer can generate non-physical intermediate values of $\*y$ and $\*u$, and the indirect relationship between the decision variables and the state constraints. For systems in which the model equations are expensive to solve and/or the states are highly constrained, this approach can be quite inefficient. If the forward model is replaced by a PINN, it will need to be retrained at every iteration of the optimization, as each iteration corresponds to a different operation point (different control value $\*u$ and solution $\*y$).

The second approach is a simultaneous analysis and design (SAND), which is also known as all-at-once approach or one-stage approach, in which both the controls and states comprise the decision variable space. The optimal values of the control $\*u$, solution $\*y$, and $\*\lambda$ are directly computed, avoiding possible non-physical intermediate values. The drawback of the all at once approach, when the optimality equations are solved numerically, is that it leads to huge systems of nonlinear equations coupling all the forward, adjoint, and controls at all intermediate time steps for all spatial locations. In contrast, the numerical NAND approach is more economical as it does not couple variables at different times.

The relative advantages and disadvantages of the two approaches change when ML surrogates are used instead of numerical solutions. PINNs used in a NAND approach need to be updated at each optimization iteration. Control PINNs follow the all-at-once approach, and solve the PDEs in continuous time and space, thereby avoiding the huge systems of nonlinear equations created by the traditional numerical approach. Control PINNs, due to their efficiency, make the all-at-once approach a feasible methodology for control.

%% file: sections/appendix_survey.tex
\section{Survey of current approaches in optimal control with PINNs}
\label{sec:Appendix-Survey}
Chen \textit{et} al. train an input convex recurrent neural network and subsequently solve a convex model predictive control (MPC) problem on the learned model \cite{chen2019optimal}. The main strength of this approach is the guarantee of an optimal solution, due to the convex nature of the model. The main limitation is similar to \cite{hwang2021solving}, in that they employ a two-stage framework of system identification and controller design. Success is evaluated on four different experiments conducted in the paper where the input convex recurrent neural network results are compared to that of a standard multilayer perceptron (MLP). The control action is also compared to that of the baselines of conventional optimizations. 

Antonelo \textit{et} al. introduce a new framework called Physics-Informed Neural Nets for Control (PINC) \cite{antonelo2021physicsinformed}. PINC uses data from the control action and initial state to solve an optimal control problem. One strength of this approach is the ability make predictions beyond the training time horizon for an indefinite period of time without a significant reduction in prediction capability. A limitation of this approach is offline learning the control separately from the solution operator. In other words, PINC is essentially a PINN that is amenable to being trained on the actions of an external controller, instead of learning the optimal control unsupervised. Success is evaluated through Mean Squared Error (MSE) validation error of the solution on the Van der Pol Oscillator problem.

Nellikkath and Chatzivasileiadis use the Karush-Kuhn-Tucker (KKT) conditions in training to improve the accuracy of neural networks trained for DC Optimal Power Flow  while utilizing substantially fewer data points \cite{nellikkath2021physicsinformed}. In addition, they show that such PINNs commit fewer worst-case violations than conventional neural networks.

Wang \textit{et} al. leverage physics-informed DeepONets ``as a fast and differentiable surrogate for tackling high-dimensional PDE-constrained optimization problems via gradient-based optimization in near real-time'' \cite{sifan2021deeponet}. The foundational idea behind their approach is to optimize a network that associates an outcome with a set of controllable variables. The strength of the approach comes from leveraging DeepONets in a physics-informed fashion.  This allows smaller training datasets, as their framework makes for an effective emulator. The limitation of their approach involves learning the control separately from the system state. The paper proposes sequentially training a neural network to learn the solution operator of a given PDE system, and then passing that information to another neural network to learn to associate the input system state with a certain control action. This approach, like others mentioned previously, is a two-stage framework. Moreover, the challenge of incorporating adjoint equations into their framework is circumvented,  where indeed there is empirical evidence in favor of using adjoint information \cite{chennault2021sgml}. Here, one measure of performance considered is training time of PINNs compared to that of traditional numerical solvers. This benchmarking is conducted for both optimal control of heat transfer, and drag minimization of obstacles in Stokes flow. Moreover, a numerical solver is utilized to validate and test the inferred control solution versus the found solution.

Hwang \textit{et} al. propose a two-stage framework for solving PDE-constrained control problems using operator learning \cite{hwang2021solving}. They first train an autoencoder model, and then infer the optimal control by fixing the learnable parameter and minimizing their objective function. One strength of their approach is the ability to apply their framework to both data-driven and data-free cases. The main downside to their approach is the two-stage nature of the framework, as the control is found only after a surrogate model has been trained. Success is measured through tracking the relative error against numerical simulation in the case of data-driven experiments, and the relative error against an analytical solution in the case of data-free experiments. Additionally, visual inspection of the trained solution operators is conducted.

Mowlavi and Nabi conduct an evaluation of the comparative performance between traditional PINNs and classic direct-adjoint-looping (DAL) to solve optimal control problems \cite{mowlavi2021optimal}. Their optimal control problem is separated into two subproblems. At each state of the system, the PDE is solved with one neural network. That information is then used by another neural network to solve for the optimal control at that given state of the PDE. Afterwards, the adjoint PDE is solved in backwards time.
The strength of Mowlavi and Nabi's approach is providing an evaluative comparison between PINNs and DAL frameworks for solving PDE-constrained optimal control problems. Their comparison provides a frame of reference for PINNs. Success is measured via validation and evaluation steps. Validation is done by monitoring residual, boundary, and initial loss components with a known solution. Evaluation is done by comparing the control cost objective with a solution found by a high-fidelity numerical solver.
One limitation of this approach is using the more easily solvable steady state Navier-Stokes, instead of unsteady state Navier-Stokes. Moreover, manual derivation is used in their DAL approach, which is unnecessary because DAL can use automatic differentiation (AD). Consequently, to a certain extent, the optimal control problem is being solved manually. Furthermore, the control of the system is being dampened over time. This is suspicious, as it might mean this dampening approach was added post hoc because of struggling results. It should be noted that the adjoint PDE is not being solved in their cost function, and thus the respective adjoint formulas are not present in said cost function. This brings us to the the main contributions of this paper.

The approaches in \cite{hwang2021solving,chen2019optimal,antonelo2021physicsinformed, mowlavi2021optimal, sifan2021deeponet} build accurate PINNs and use them as differentiable surrogate models in the optimization solution for control applications.
Going beyond these prior approaches, we propose a framework that can be considered a new taxonomical entity within the genus of PINNs, wherein the PDE-constrained optimization is fused with the training loss function to create a one-stage approach to directly learning both the system solution and the optimal control.

%% file: sections/methodology.tex
\section{Methodology}
\label{sec:Methodology}

\begin{algorithm}[t]
\SetAlgoLined
\KwResult{Training of a Control PINN that learns the optimal solution and the optimal control function for the given problem in \eqref{eq:pde-control-problem}}
 {
    \begin{enumerate}[leftmargin=16pt]
        \item Construct a network with inputs $t$, $x$, and outputs $\*y$, $\*u$, and $\*\lambda$  based on the  architecture in \cref{fig:nn_architecture}.
        \item Compute the necessary derivatives of the outputs w.r.t to the inputs and other outputs via backpropagation. These derivatives will be used to form the residuals in the loss function \cref{eqn:loss-function-extended}.
        \item With $\*y_*^i$ denoting the training data for the input $(t^i,x^i)$, and $\*y^i , \*u^i ,\*\lambda^i$  corresponding outputs of the network,  $\*\theta_k$ representing the weights of the network at iteration $k$, and $\Vert \cdot \Vert$ as $L_2$ norm, we seek to minimize the loss function:
        \vspace{-0.5em}
        \begin{equation}
        \label{eqn:loss-function-extended}
        \begin{split}
        L\left({\*\theta_k}\right) &= \sum_{i} \,\Vert \*y_*^i - \*y^i \Vert^2\\
        & + \sum_{i} \, \Vert \pdv{\*y^i}{t} - f\left(\*y^i,\*u^i\right) \Vert^2 \\
        & + \sum_{i} \, \Vert  \pdv{\*\lambda^i}{t} + \*(\*\lambda^i)^T \pdv{\*f}{\*y^i} \left(\*y^i,\*u^i\right) + \pdv{g}{\*y^i} \left(\*y^i,\*u^i\right) \Vert^2 \\
        & + \sum_{i} \, \Vert \left(\*\lambda^i\right)^T \pdv{\*f}{\*u^i} \left(\*y^i,\*u^i\right) + \pdv{g}{\*u^i} \left(\*y^i,\*u^i\right) \Vert^2.
        \end{split}
        \end{equation}
    \item Update the weights of the network using the optimizer according to the loss function \cref{eqn:loss-function-extended}.
    \vspace{-0.5em}
    $$\*\theta_{k+1} = \texttt{ADAM}(\*\theta_{k},\grad_{\*\theta} L\left({\*\theta_k}\right)).$$
    \item Repeat until convergence.
    \end{enumerate} 
 }
 \caption{The procedure to train a Control PINN model.}
 \label{algorithm:control-pinn}
\end{algorithm}

\input{tikz/control-pinn}


  

Hairer and Wanner provide an overview of optimal control problems \cite[pp. 461-463]{Hairer2002SolvingOD}. 
Consider the following PDE-constrained control problem:
\begin{equation}
\label{eq:pde-control-problem}
\begin{gathered}
\*u^* = \argmin_{\*u} \Psi(\*u)  = \int_{t_0}^{t_f} \int_\Omega g\left(\*y,\*u\right)  \dd{x} \dd{t} +  \int_\Omega w({\*y}|_{t_f})  \dd{x}
+ \int_{t_0}^{t_f} \int_{\partial\Omega} h\left(\*y,\*u\right)  \dd{x} \dd{t} ,  \\
\textnormal{subject to the PDE:}\quad
\begin{cases}
\pdv{\*y }{t} = \*f\left(\*y,\*u\right),& \forall  t \in [t_0,t_f],~~  \forall x \in \Omega,\\
\*y|_{t_0} = \*y_0,\,& \forall x \in \Omega,\\
\!B\,\*y  = \*b(t, x),\,& \forall  t \in [t_0,t_f],~~  \forall x \in \partial\Omega,
\end{cases}
\end{gathered}
\end{equation}
%
where $\*y = \*y(t,x)$ is time and space-dependant system state and $\*u = \*u(t,x)$ is a time-varying distributed control signal applied to the system. $\*f$ is the right-hand-side equation of the PDE that describes the dynamics of the system in terms of $\*y$ and its spatial derivatives. $\!B$ and $\*b$ define the boundary conditions, $\*y_0$ is the initial condition of the system state,  $g$ is a cost over the spatio-temporal domain, $w$ is a cost at the final time, and $h$ is a cost over the boundary. In this work we assume that the $h$ term is zero everywhere. The functions $\*y$ and $\*u$ live in appropriate function spaces with sufficient smoothness \cite[Sec. 1.5]{Hinze2009}. We are interested in finding optimal solution-control pairs $(\*y^*,\*u^*)$ that satisfy the PDE while minimizing a given cost functional  $\Psi $ that may depend on the solution, control, or both.

One way  to solve~\cref{eq:pde-control-problem} is through its Lagrangian, which we write as,
\begin{align}
\label{eqn:opt-problem-Lag}
\mathfrak{L} &= \int_{t_0}^{t_f}\int_\Omega \left[ g\left(\*y,\*u\right) - {\*\lambda}^T \left(\pdv{\*y}{t} - f(\*y,\*u)\right) \right] \dd{x}\dd{t} + \int_\Omega w({\*y}|_{t_f})\dd{x},
\end{align}
in terms of the adjoint variable $\*\lambda(t,x)$. Under first order optimality conditions~\cite{wright1999numerical}, a local minimum of~\cref{eqn:opt-problem-Lag} can be attained when,
\begin{subequations}\label{eqn:point-wise-equations-optim-control}
\begin{align}
    \begin{split}
        &\pdv{\*y }{t} = \*f\left(\*y,\*u\right),~~ \forall  t \in [t_0,t_f],~~  \forall x \in \Omega,\label{eqn:fwd-pde}\\
        &{\*y}|_{t_0}  = {\*y_0}, ~~ \forall x \in \Omega,\\
        &\!B\,\*y  = \*b(t, x),~~ \forall  t \in [t_0,t_f],~~  \forall x \in \partial\Omega,
    \end{split}\\
    \begin{split}
        &\pdv{\*\lambda }{t} = - {\*\lambda}^T \pdv{\*f}{\*y}\left(\*y,\*u\right)- \pdv{g}{\*y}\left(\*y,\*u\right), ~~ \forall  t \in [t_0,t_f],~~  \forall x \in \Omega,\\
        &{\*\lambda}|_{t_f} =  w_{\*y}({\*y}|_{t_f}), ~~\forall x \in \Omega,  \label{eqn:bkwd-pde} \\
        &{\!B^*} {\*\lambda}  = 0, ~~ \forall t \in [t_0,t_f],~~  \forall x \in \Omega,
    \end{split}\\
    & \*\lambda ^T \pdv{\*f}{\*u}\left(\*y,\*u\right) + \pdv{g}{\*u}\left(\*y,\*u\right)= \*0 ,~~ \forall t \in [t_0,t_f],  ~~\forall x \in \Omega \label{eqn:system-push-back},
\end{align}
\end{subequations}
%

are satisfied point-wise over time $t$ and space $x$, where \cref{eqn:fwd-pde} represents the evolution of the system state, \cref{eqn:bkwd-pde} represents the evolution of the adjoint equation, and \cref{eqn:system-push-back} represents the optimality equation. Computing the adjoint variable $\*\lambda$ is necessary because all of the variables are coupled in the optimality conditions \eqref{eqn:point-wise-equations-optim-control}. Additionally, the adjoint $\*\lambda$ provides the sensitivity of the loss function with respect to changes in the constraints \cite{wright1999numerical} which can be used for diagnostics and system optimization, and its Control PINN evaluation requires less computational effort than using directional derivatives \cite[Sec. 1.6.2]{Hinze2009}.
As an example, in the context of an autonomous vehicles, the system state (the velocity and the position of the vehicle) are determined by the equations of motion. By the same token, the system controller would be the software that governs the steering wheel, acceleration, and braking. Finally, the adjoint (or co-state variable) would be the response of the vehicle to the software's choices of direction and speed.

\begin{wrapfigure}{r}{0.7\textwidth}
  \centering
  \includegraphics[width=0.7\textwidth]{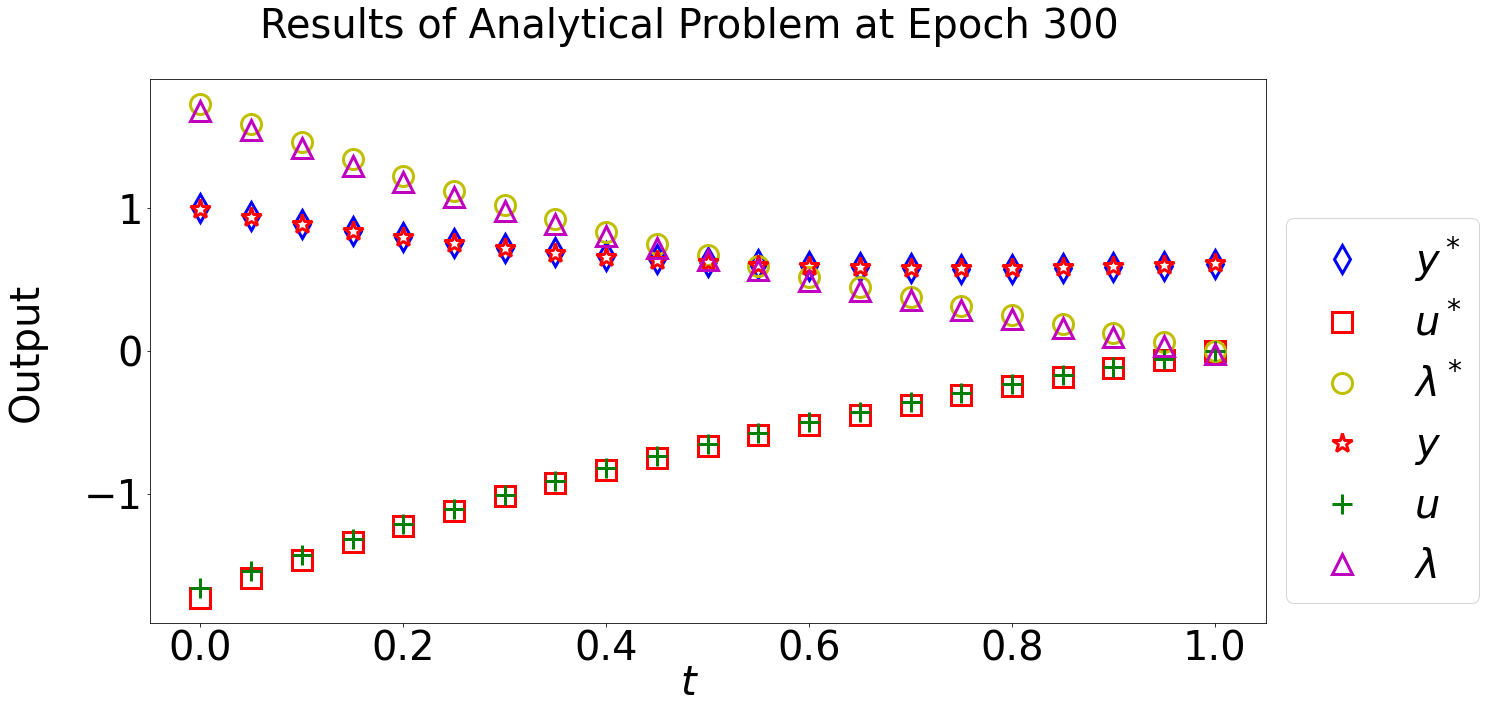}
  \caption{\textmd{\emph{Results of the analytical problem:} The optimal solutions of the system state, control, and adjoint on the control are respectively denoted by $\*y^*$, $\*u^*$, and $\*\lambda^*$. Their corresponding learned solutions found by the Control PINN model are denoted by $\*y$, $\*u$, and $\*\lambda$. After 300 epochs the Control PINN has reached convergence on the analytical solution, and the solutions remain unchanged upon further training.}}
  \label{fig:analytical-results}
\end{wrapfigure}

We design a neural network that generates triples $( \*y, \*u, \*\lambda )$ from input data  $(t,x)$, equipped with PINN loss functions that capture the first order optimality conditions according to \cref{eqn:point-wise-equations-optim-control}. 
The process of solving the control problem \cref{eq:pde-control-problem} is outlined in \cref{algorithm:control-pinn}. 
The main technical challenge involves learning the state of a dynamical system while at the same time finding its optimal control. We create a path in the computational graph for the backpropagation of derivatives by placing the the deep layers that generate $\*u$ after the layers that generate $\*y$ and, similarly for $\*\lambda$. This ensures that all necessary derivatives for \cref{eqn:point-wise-equations-optim-control} can be computed using automatic differentiation.

One limitation would be approximating chaotic systems over long time intervals. However, any given PINN faces this difficulty. Similarly, finding optimal controls for chaotic systems is difficult even for standard numerical methods. It should be noted that the adjoints of chaotic atmospheric dynamics are used successfully at the European Centre for Medium-Range Weather Forecasts (ECMWF) for data assimilation in numerical weather prediction. Additionally, of note is the work of Qiqi Wang and collaborators in alternative definitions of adjoints for chaotic systems \cite{Blonigan2017ChaoticAdjoint}. The scope of Control PINN is mainly concerned with controls for non-chaotic systems, as is typical in engineering systems.

There is a tension between different terms in the loss function defined in \cref{algorithm:control-pinn}. For example, satisfying the boundary conditions may oppose adhering to the constraints imposed by the physical laws. This is addressed by scaling factors in the loss function and treating them as hyperparameters. We further note that as Control PINN is applied to increasingly more complex problems, a unique solution to the problem may not be available. We discuss the validation of the solution found by the Control PINN model as well as its optimality further in \cref{sec:1d-heat-equation}.

\textbf{Analysis of convergence:} With the overlap of fields of study comes an overlap of definitions. In numerical analysis, the term \emph{convergence} refers to the behavior of numerical solutions as the discretization time step or grid size becomes small. In PINNs, there is no counterpart of a small parameter going to zero. 
Perhaps the convergence of a PINN could be observed via studying approximation errors as its dimension is increased.
While the universal approximation property of NNs \cite{Hornik1989UniversalApproximators} has long ago been established, \emph{convergence-type} results wherein the approximation error is bounded by a function of the number of neurons is not yet generally available.
Extending the concept of ``convergence” to encompass PINNs is important to link the worlds of machine learning and numerical mathematics. Such an endeavor requires considerable additional investigation, and is beyond the scope of this paper.

\textbf{Control PINN architecture:} A visual representation of the Control PINN architecture is found in \cref{fig:nn_architecture}. Adaptive moment estimation (ADAM) is used as the optimizer. The activation function of exponential linear unit (ELU) is used. The neural density is 100 neurons per layer.
There are five hidden layers that proceed the input layer that takes in time ($t$) and space ($x$). 
The information of the system state ($\*y$) is passed to the controller ($\*u$) both directly and indirectly by a skip connection and three hidden layers, respectively. The aggregate information of the system state ($\*y$) and controller ($\*u$) is handled similarly in the context of the system's push back on the controller ($\*\lambda$), except with only two hidden layers. This architecture enables for the automatic differentiation of second order and mixed derivatives. This is necessary to impose the custom loss function detailed in \cref{algorithm:control-pinn}.

\textbf{Choice of optimizer:} ADAM is used as the optimizer because at each iteration of training the sampling points are randomized. Resampling of random points is very important to ensure that the PINN finds a nontrivial solution. If static sampling points had been used, then Limited-memory Broyden-Fletcher-Goldfarb-Shanno (L-BFGS) would be appropriate. However, using a fixed set of collocation points can be detrimental to the convergence of PINNs to the correct solution \cite{Daw2022ImportanceOfSampling}. Thus, L-BFGS should not be used as the optimizer. That being said, it is worth investigating whether the use of evolutionary sampling (incrementally accumulating collocation points in regions of high PDE residuals) could improve the accuracy of the learned solution by Control PINN \cite{Daw2022ImportanceOfSampling}.

\textbf{Regularization of weights:} The current implementation does not scale boundary conditions, but scales the loss components by a constant ($10^{-1}$). By regularizing the weights in this way, the scale of each loss component is brought in line to be proportional to one another. In additional experiments not detailed in this paper, constant scaling of the loss components with values less than $1\text{e-}1$ resulted in poor performance. Thus, the relative scaling of each loss components' regularized weight was determined using a hand-tuning approach. To justify the choice of constant scaling, and possibly improve upon the accuracy of the learned solutions, future work will include an ablation study of the following: (1) A learning rate annealing algorithm that leverages gradient statistics to balance the interaction between loss components \cite{Wang2020gradientpath}. (2) Curriculum regularization, in which Control PINN's loss term could start from a simple PDE regularization, and increase in complexity as training progresses \cite{Krishnapriyan2021failuremodes}. (3) Augmented Lagrangian relaxation method for PINNs to adaptively balance each loss component \cite{Son2022AugLagPINN}.

\textbf{Hyperparameters:} In order to minimize the effects of hyperparameters, the network depth, neural density per layer, and overall architecture are kept exactly the same for each of the three problems presented in this paper. With all three increasingly challenging experiments presented in this paper using the same architecture, the robustness of the Control PINN framework is demonstrated in its ability to tackle each one without the need for hyperparameter tuning adjustments via random search or grid search.

\biblio

%% file: tikz/control-pinn.tex
\tikzset{every picture/.style={line width=0.3pt}} 
\begin{figure}[t]
\centering

\begin{tikzpicture}[x=0.75pt,y=0.75pt,yscale=-0.8,xscale=0.8]

\draw   (10.29,181.06) -- (42.92,181.06) -- (42.92,210.26) -- (10.29,210.26) -- cycle ;
\draw    (42.92,194.74) -- (79,194.74) ;
\draw [shift={(81,194.74)}, rotate = 180] [color={rgb, 255:red, 0; green, 0; blue, 0 }  ][line width=0.75]    (10.93,-3.29) .. controls (6.95,-1.4) and (3.31,-0.3) .. (0,0) .. controls (3.31,0.3) and (6.95,1.4) .. (10.93,3.29)   ;
\draw   (10.29,233.53) -- (42.92,233.53) -- (42.92,262.72) -- (10.29,262.72) -- cycle ;
\draw    (42.92,249.05) -- (79,249.05) ;
\draw [shift={(81,249.05)}, rotate = 180] [color={rgb, 255:red, 0; green, 0; blue, 0 }  ][line width=0.75]    (10.93,-3.29) .. controls (6.95,-1.4) and (3.31,-0.3) .. (0,0) .. controls (3.31,0.3) and (6.95,1.4) .. (10.93,3.29)   ;
\draw   (173.14,210.91) -- (205.77,210.91) -- (205.77,240.1) -- (173.14,240.1) -- cycle ;
\draw   (249.88,206.75) .. controls (252.85,212.98) and (255.7,218.9) .. (259,218.9) .. controls (262.3,218.9) and (265.15,212.98) .. (268.12,206.75) .. controls (271.1,200.52) and (273.95,194.59) .. (277.25,194.59) .. controls (280.55,194.59) and (283.4,200.52) .. (286.37,206.75) .. controls (288.63,211.48) and (290.82,216.03) .. (293.18,217.95) ;
\draw   (250.36,223.82) .. controls (253.34,230.04) and (256.18,235.97) .. (259.49,235.97) .. controls (262.79,235.97) and (265.63,230.04) .. (268.61,223.82) .. controls (271.59,217.59) and (274.43,211.66) .. (277.73,211.66) .. controls (281.04,211.66) and (283.88,217.59) .. (286.86,223.82) .. controls (289.12,228.55) and (291.3,233.1) .. (293.67,235.02) ;
\draw   (250.85,242.96) .. controls (253.82,249.18) and (256.67,255.11) .. (259.97,255.11) .. controls (263.27,255.11) and (266.12,249.18) .. (269.1,242.96) .. controls (272.07,236.73) and (274.92,230.8) .. (278.22,230.8) .. controls (281.52,230.8) and (284.37,236.73) .. (287.34,242.96) .. controls (289.6,247.68) and (291.79,252.24) .. (294.16,254.16) ;
\draw   (242.48,187.6) -- (300.81,187.6) -- (300.81,263.96) -- (242.48,263.96) -- cycle ;

\draw    (137.52,226.39) -- (170.43,226.39) ;
\draw [shift={(172.43,226.39)}, rotate = 180] [color={rgb, 255:red, 0; green, 0; blue, 0 }  ][line width=0.75]    (10.93,-3.29) .. controls (6.95,-1.4) and (3.31,-0.3) .. (0,0) .. controls (3.31,0.3) and (6.95,1.4) .. (10.93,3.29)   ;
\draw    (207.01,226.39) -- (239.92,226.39) ;
\draw [shift={(241.92,226.39)}, rotate = 180] [color={rgb, 255:red, 0; green, 0; blue, 0 }  ][line width=0.75]    (10.93,-3.29) .. controls (6.95,-1.4) and (3.31,-0.3) .. (0,0) .. controls (3.31,0.3) and (6.95,1.4) .. (10.93,3.29)   ;
\draw   (336.64,210.91) -- (369.27,210.91) -- (369.27,240.1) -- (336.64,240.1) -- cycle ;
\draw   (413.37,206.75) .. controls (416.35,212.98) and (419.2,218.9) .. (422.5,218.9) .. controls (425.8,218.9) and (428.65,212.98) .. (431.62,206.75) .. controls (434.6,200.52) and (437.44,194.59) .. (440.75,194.59) .. controls (444.05,194.59) and (446.89,200.52) .. (449.87,206.75) .. controls (452.13,211.48) and (454.31,216.03) .. (456.68,217.95) ;
\draw   (413.86,223.82) .. controls (416.84,230.04) and (419.68,235.97) .. (422.98,235.97) .. controls (426.29,235.97) and (429.13,230.04) .. (432.11,223.82) .. controls (435.08,217.59) and (437.93,211.66) .. (441.23,211.66) .. controls (444.53,211.66) and (447.38,217.59) .. (450.36,223.82) .. controls (452.62,228.55) and (454.8,233.1) .. (457.17,235.02) ;
\draw   (414.35,242.96) .. controls (417.32,249.18) and (420.17,255.11) .. (423.47,255.11) .. controls (426.77,255.11) and (429.62,249.18) .. (432.59,242.96) .. controls (435.57,236.73) and (438.42,230.8) .. (441.72,230.8) .. controls (445.02,230.8) and (447.87,236.73) .. (450.84,242.96) .. controls (453.1,247.68) and (455.29,252.24) .. (457.65,254.16) ;
\draw   (405.98,187.6) -- (464.31,187.6) -- (464.31,263.96) -- (405.98,263.96) -- cycle ;

\draw    (301.02,225.77) -- (333.93,225.77) ;
\draw [shift={(335.93,225.77)}, rotate = 180] [color={rgb, 255:red, 0; green, 0; blue, 0 }  ][line width=0.75]    (10.93,-3.29) .. controls (6.95,-1.4) and (3.31,-0.3) .. (0,0) .. controls (3.31,0.3) and (6.95,1.4) .. (10.93,3.29)   ;
\draw    (370.5,225.77) -- (403.42,225.77) ;
\draw [shift={(405.42,225.77)}, rotate = 180] [color={rgb, 255:red, 0; green, 0; blue, 0 }  ][line width=0.75]    (10.93,-3.29) .. controls (6.95,-1.4) and (3.31,-0.3) .. (0,0) .. controls (3.31,0.3) and (6.95,1.4) .. (10.93,3.29)   ;
\draw   (500.62,210.91) -- (533.26,210.91) -- (533.26,240.1) -- (500.62,240.1) -- cycle ;
\draw    (465,225.51) -- (497.92,225.51) ;
\draw [shift={(499.92,225.51)}, rotate = 180] [color={rgb, 255:red, 0; green, 0; blue, 0 }  ][line width=0.75]    (10.93,-3.29) .. controls (6.95,-1.4) and (3.31,-0.3) .. (0,0) .. controls (3.31,0.3) and (6.95,1.4) .. (10.93,3.29)   ;
\draw  [fill={rgb, 255:red, 255; green, 255; blue, 255 }  ,fill opacity=1 ] (222,71.42) -- (222,42) -- (364.34,42) .. controls (364.34,60.39) and (415.67,56.71) .. (382.46,71.42) -- cycle ; \draw  [fill={rgb, 255:red, 255; green, 255; blue, 255 }  ,fill opacity=1 ] (243.57,75.09) -- (243.57,45.68) -- (385.91,45.68) .. controls (385.91,64.06) and (437.24,60.39) .. (404.02,75.09) -- cycle ; \draw  [fill={rgb, 255:red, 255; green, 255; blue, 255 }  ,fill opacity=1 ] (265.13,78.77) -- (265.13,49.35) -- (407.47,49.35) .. controls (407.47,67.74) and (458.8,64.06) .. (425.59,78.77) -- cycle ;
\draw  [dash pattern={on 4.5pt off 4.5pt}]  (190,132) -- (190,212) ;
\draw  [dash pattern={on 4.5pt off 4.5pt}]  (190,132) -- (276,132) ;
\draw  [dash pattern={on 4.5pt off 4.5pt}]  (276,132) -- (276,83.67) ;
\draw [shift={(276,81.67)}, rotate = 90] [color={rgb, 255:red, 0; green, 0; blue, 0 }  ][line width=0.75]    (10.93,-3.29) .. controls (6.95,-1.4) and (3.31,-0.3) .. (0,0) .. controls (3.31,0.3) and (6.95,1.4) .. (10.93,3.29)   ;
\draw  [dash pattern={on 4.5pt off 4.5pt}]  (351,211) -- (351,84.33) ;
\draw [shift={(351,82.33)}, rotate = 90] [color={rgb, 255:red, 0; green, 0; blue, 0 }  ][line width=0.75]    (10.93,-3.29) .. controls (6.95,-1.4) and (3.31,-0.3) .. (0,0) .. controls (3.31,0.3) and (6.95,1.4) .. (10.93,3.29)   ;
\draw  [dash pattern={on 4.5pt off 4.5pt}]  (519.33,130) -- (519.33,210) ;
\draw  [dash pattern={on 4.5pt off 4.5pt}]  (519.33,130) -- (417.67,129.33) ;
\draw  [dash pattern={on 4.5pt off 4.5pt}]  (417.67,129.33) -- (417.67,83) ;
\draw [shift={(417.67,81)}, rotate = 90] [color={rgb, 255:red, 0; green, 0; blue, 0 }  ][line width=0.75]    (10.93,-3.29) .. controls (6.95,-1.4) and (3.31,-0.3) .. (0,0) .. controls (3.31,0.3) and (6.95,1.4) .. (10.93,3.29)   ;
\draw    (189.33,241) -- (189.33,314) ;
\draw   (88.88,206.75) .. controls (91.85,212.98) and (94.7,218.9) .. (98,218.9) .. controls (101.3,218.9) and (104.15,212.98) .. (107.12,206.75) .. controls (110.1,200.52) and (112.95,194.59) .. (116.25,194.59) .. controls (119.55,194.59) and (122.4,200.52) .. (125.37,206.75) .. controls (127.63,211.48) and (129.82,216.03) .. (132.18,217.95) ;
\draw   (89.36,223.82) .. controls (92.34,230.04) and (95.18,235.97) .. (98.49,235.97) .. controls (101.79,235.97) and (104.63,230.04) .. (107.61,223.82) .. controls (110.59,217.59) and (113.43,211.66) .. (116.73,211.66) .. controls (120.04,211.66) and (122.88,217.59) .. (125.86,223.82) .. controls (128.12,228.55) and (130.3,233.1) .. (132.67,235.02) ;
\draw   (89.85,242.96) .. controls (92.82,249.18) and (95.67,255.11) .. (98.97,255.11) .. controls (102.27,255.11) and (105.12,249.18) .. (108.1,242.96) .. controls (111.07,236.73) and (113.92,230.8) .. (117.22,230.8) .. controls (120.52,230.8) and (123.37,236.73) .. (126.34,242.96) .. controls (128.6,247.68) and (130.79,252.24) .. (133.16,254.16) ;
\draw   (81.48,187.6) -- (139.81,187.6) -- (139.81,263.96) -- (81.48,263.96) -- cycle ;

\draw    (189.33,314) -- (434.33,314) ;
\draw    (434.33,314) -- (434.33,267) ;
\draw [shift={(434.33,265)}, rotate = 90] [color={rgb, 255:red, 0; green, 0; blue, 0 }  ][line width=0.75]    (10.93,-3.29) .. controls (6.95,-1.4) and (3.31,-0.3) .. (0,0) .. controls (3.31,0.3) and (6.95,1.4) .. (10.93,3.29)   ;

\draw (20.53,187.58) node [anchor=north west][inner sep=0.75pt]    {$t$};
\draw (19.07,242.5) node [anchor=north west][inner sep=0.75pt]    {$x$};
\draw (182.47,215.14) node [anchor=north west][inner sep=0.75pt]    {$\*y$};
\draw (345.03,216.14) node [anchor=north west][inner sep=0.75pt]    {$\*u$};
\draw (509.32,213.74) node [anchor=north west][inner sep=0.75pt]    {$\*\lambda $};
\draw (314.77,51.8) node [anchor=north west][inner sep=0.75pt]    {$[ \*y,\*u,\*\lambda ]$};

\end{tikzpicture}
    \caption{\textmd{\emph{Control PINN architecture:} Time $t$ and space $x$ are inputs to the model, whereas the system state $\*y$, control $\*u$, and adjoint $\*\lambda$ are the outputs to the model. The boxes with wave lines represent hidden layers. The dashed lines indicate the outputs. Note that the system state $\*y$ is passed into both the control $\*u$ and the adjoint $\*\lambda$. This enables for the automatic differentiation of second order and mixed derivatives. Additionally, please note there is a nonlinear connection between the inputs $t$ and $x$ and the outputs $\*u$ and $\*\lambda$, making both outputs functions of time and space.}}
    \label{fig:nn_architecture}
\end{figure}

%% file: sections/analytical_problem.tex
\section{Analytical problem}
\label{sec:analytical-toy-problem}

As a proof of concept, and to provide a foundation for methodology validation, consider the following ODE control problem \cite[Ch. 6, pp. 272-273, eqn. 65]{Hager2000RungeKuttaMI},
\begin{equation}
\label{eqn:toy-problem}
\begin{split}
& \*u^* = \argmin_{u} \Psi(u)  = \int_{t_0}^{t_f} \left( \*y(t)^2 + \frac{1}{2} \, \*u(t)^2 \right) \dd{t}, \quad  \\
& \textnormal{subject to}\quad
\begin{cases}
\pdv{\*y}{t}(t) = \frac{1}{2} \*y + \*u,\,& \forall  t \in [t_0,t_f],\\
\*y(t_0) = \*y_0 = 1,\\
\pdv{\*\lambda}{t}(t) = - \frac{1}{2}\, \*\lambda(t) - 2 \*y,& \forall t \in [t_f,t_0],\\
\*\lambda(t_f) = 0, &\\
0 = -\*\lambda(t) - \*u,& \forall t \in [t_0,t_f],
\end{cases}
\end{split}
\end{equation}
where $t_0 = 0$, and $t_f = 1$. An analytical form of the optimal solution is known:
\begin{equation}
\begin{aligned}
    \*y^*(t) &= \frac{2 e^{3t} + e^3}{e^{3t/2} (2+e^3)},\quad \*u^*(t) = \frac{2 (e^{3t} - e^3)}{e^{3t/2} (2+e^3)},\quad \*\lambda^*(t) = -\*u^*(t).
\end{aligned}
\end{equation}

After 300 epochs of training, the Control PINN converges on the optimal solution. This is shown in \cref{fig:analytical-results}, wherein $\*y^*$, $\*u^*$, and $\*\lambda^*$ respectively represent the reference solution for the system state $y$, the reference solution for the system controller $u$, and the reference solution of the adjoint on the controller $\*\lambda$. An animation of the convergence of the outputs to the reference solution as training progresses can found at: 
\url{https://github.com/ComputationalScienceLaboratory/control-pinns}.

\biblio

%% file: sections/heat_equation.tex
\section{One-dimensional heat equation}
\label{sec:1d-heat-equation}
  
Fourier's famous heat equation $\frac{\partial u}{\partial t} = \alpha^2 \frac{\partial^2 u}{\partial x^2}$ is the origin of the Fourier series theory \cite[p. 30]{Hairer2002SolvingOD}, and is thus a well known and established problem with which to illustrate the robustness of Control PINN. Imagine an infinitesimally thin steel beam being heated by a heat pad. This heat pad will be controlled in heating the steel beam such that a given temperature will be reached at the final time. The problem set up is as follows:
\begin{equation}
\label{eqn:1d-heat}
\begin{split}
\*u^* = \argmin_{\*u} \Psi(\*u)  &= 
\int_{\Omega} {\big(} \left( \*y(t_0,x) - \*y^*(t_0,x) \right){\big)}^2 \, \mathrm{d}x  \\
&  + \int_{\Omega} \left( \*y(t_f,x) - \*y^*(t_f,x) \right)^2 \, \mathrm{d}x  \\ 
& + \int_{\Omega} \int_{t_0}^{t_f}   \left({\*u(t,x)}\right) ^2 \, \mathrm{d}t \mathrm{d}x,  \\
& \textnormal{subject to}\quad
\begin{aligned}
\pdv{\*y}{t}  &=  0.1 \pdv[2]{\*y}{x} + \*u(t,x), && \forall  t \in [t_0,t_f],  ~~x \in \Omega,  \\
\*y(t_0,x) &= \sin{(\pi x)} \sin{(2 \pi x)},&& \forall x \in \Omega,\\
\*y(t,x) &= 0,&& \forall t \in [t_0, t_f], \, x \in \Omega, 
\end{aligned}\\
\*y^*(t,x) &= \frac{2}{\pi + 4 \pi^3} \left(e^{-\pi^2 t} - \cos{\frac{\pi t}{2}}  + 2 \pi \sin{\frac{\pi t}{2}}\right) \sin{\pi x},
\end{split}
\end{equation}
where $t_0 = 0$, $t_f=1$ and $\Omega = [0, 1]$. The first term in the cost in~\cref{eqn:1d-heat} represents the PINN loss, ensuring that the learned equation obeys the initial conditions, the second term represents the desired final distribution of heat along the beam, and the third term represents our desire to minimize the energy used to control the temperature.

\begin{figure}[t!]
    \centering
    \includegraphics[width=0.65\textwidth]{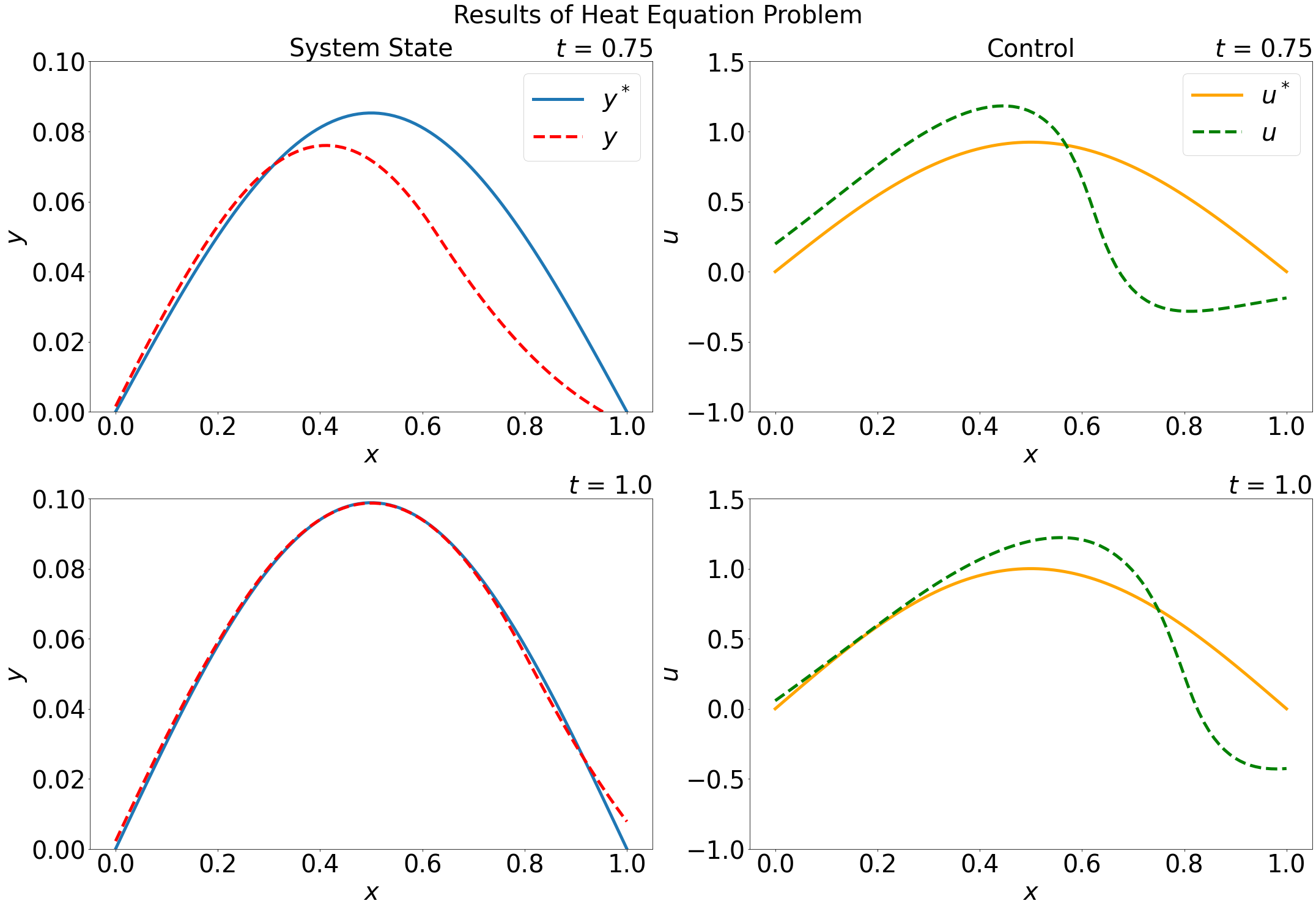}
    \caption{\textmd{\emph{One-dimensional heat equation results:} \emph{Left:} Comparison of the learned system state $\*y$ and the reference system state $\*y^*$. \emph{Right:} Comparison of the learned control $\*u$ and the reference control $\*u^* = sin(\pi x) sin(\frac{\pi t}{2})$. Note that there are multiple control signals that satisfy the PDE solution. The reference control $\*u^*$ might not be the optimal solution, and is not the same as the control signal found by Control PINN. Control PINN found a different control that minimizes the control action on the system state. Please note that the control signal $\*u$ can be negative, and would thus represent cooling of the system.
    }}
    \label{fig:1d-heat-results}
\end{figure}

\Cref{fig:1d-heat-results} illustrates the results of the model finding the solutions of the system state and the control over the time span $[0,1]$. A reference solution for the adjoint $\*\lambda$ is not known, and is therefore not included in the plot for comparative purposes. There is an exact match between the reference solution and learned solution of the system state.

Of note is the shape of the optimal control found by the model. The optimal control of the reference solution reaches a parabolic curve over the domain  as time approaches $t = 1$. The trajectory of the model's solution dips more sharply downward towards the end of the feature space. This begs the question of if the model's solution of the control is valid.

\begin{wraptable}{r}{0.35\textwidth}
    \caption{\textmd{\emph{One-dimensional heat equation:} Relative error of the the system states between the reference solution and the solution computed from a numerical solver using Control PINN control data.}}
    \label{table:heat-equation-relative-error}
    \begin{centering}
    \begin{tabular}{cc}
        \toprule
        Time & Relative Error\\
        \midrule
        0.1 & 1.2309\\
        0.2 & 0.2646\\
        0.3 & 0.1805\\
        0.4 & 0.2010\\
        0.5 & 0.1843\\
        0.6 & 0.1514\\
        0.7 & 0.1168\\
        0.8 & 0.0843\\
        0.9 & 0.0535\\
        1.0 & 0.0232\\
    \bottomrule
    \end{tabular}
    \end{centering}
\end{wraptable}

To validate the solution found by the Control PINN, we generated high resolution offline data of the control and the solution outputs at different values of time and space. A direct numerical solver (DNS) using finite differences was then used along with the control data to generate DNS solutions for this problem. The reference solution of the system control is denoted by $\*u^* = \sin(\pi x) \sin(\frac{\pi t}{2})$. We compute the mean squared integral of the control $\*u$ for both the Control PINN and the analytical solution, and we find that Control PINN is slightly better.

\begin{figure}[t!]
    \centering
    \begin{subfigure}[t]{0.45\textwidth}
        \centering
        \includegraphics[width=\textwidth]{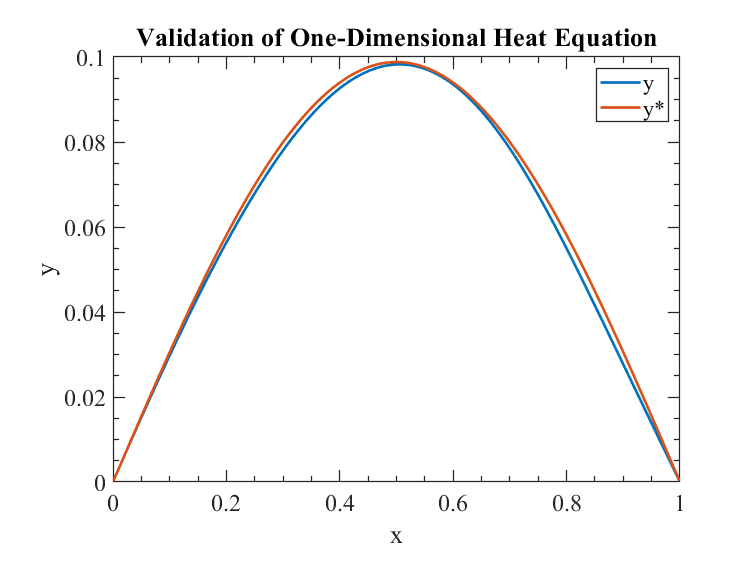}
    \end{subfigure}
    ~
    \begin{subfigure}[t]{0.45\textwidth}
        \centering
        \includegraphics[width=\textwidth]{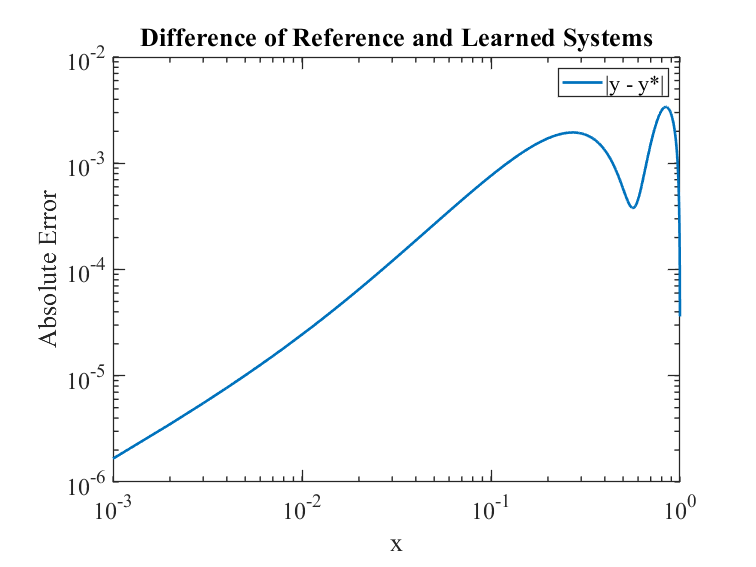}
    \end{subfigure}
    \caption{\textmd{\emph{Validation of 1-D Heat Equation:} 
    High resolution offline control data from Control PINN was used in a numerical solver to compute the system state (Denoted by $\*y$). After 10,000 epochs of training to ensure a high accuracy, the mean of control action of Control PINN is slightly less than that of the reference control (0.2475 and 0.2497, respectively).
    \emph{Left:} 
    Comparison of the system state between the reference solution and the solution computed by the numerical solver using Control PINN control data ($\*y^*$ and $\*y$, respectively).
    \emph{Right:} 
    On a log-log scale, the absolute error between $\*y^*$ and $\*y$ is shown.
    }}
    \label{fig:1d-heat-validation}
\end{figure}

\Cref{fig:1d-heat-validation} shows the validation of the solution for this problem. We see good agreement between the two approaches. \Cref{table:heat-equation-relative-error} shows relative error, computed as the $L_2$ norm, of the the system states between the reference solution and the solution computed from a numerical solver using Control PINN control data (discretized uniformly with 1,000 points) at various timepoints.

The reason for the discrepancy between $\*u^*$ and $\*u$ in \cref{fig:1d-heat-results} is that the reference solution $\*u^*$ is a solution to the PDE system, whereas the Control PINN result solves the PDE-constrained optimization involving an extra cost function on the control. Indeed, numerical validation shows that Control PINN was able to find a solution with smaller amount of control.

\biblio

%% file: sections/predator_prey.tex
\section{Two-dimensional predator-prey problem}
\label{sec:2d-predator-prey}

We next look at a two-dimensional predator-prey (Lotka-Volterra) problem formulated as a reaction-diffusion problem\cite{lotka1910reaction}.
On a two-dimensional Cartesian grid, the system characterized by  \cref{eqn:2d-predator-prey} defines the interaction of two populations: the predator and the prey. The rate of birth for each population follows an exponential law based on the current population and is also affected by the competing population.  
We are interested in controlling the prey population (over time and space) by inserting predators at different times and locations. For this example we have considered a starting profile of preys which is then ``herded'' into a desired profile over a specified timespan.  The predator population in \cref{eqn:2d-predator-prey} is denoted by $\*y_1$, whereas $\*y_2$ represents the prey. Similarly, $u_1$ and $u_2$ are predator and prey control functions, respectively. We have used $u_1(t,x) = 0$ for this problem. 
\begin{equation}
\label{eqn:2d-predator-prey}
\begin{split}
& \*u^* = \argmin_{\*u} \Psi(\*u)  = \int_{\Omega} \int_{t_0}^{t_f}  \norm{ \*y(t,x)  - \*y^*(t,x) }^2_2 \, \mathrm{d}t \, \mathrm{d}x +  \int_{\Omega} \int_{t_0}^{t_f}  \norm {\*u(t,x)} ^2_2 \, \mathrm{d}t \mathrm{d}x, \quad  \\
& \textnormal{subject to} \quad
\begin{cases}
0 = \pdv{\*y_1}{t} -  \pdv[2]{\*y_1}{x} - u_1(t,x) + \*y_1(t,x), & \forall  t \in [t_0,t_f], \forall x\in \Omega,\\
0 = \pdv{\*y_2}{t} -  \pdv[2]{\*y_2}{x} - u_2(t,x) - \*y_2(t,x), & \forall  t \in [t_0,t_f], \forall x\in \Omega,\\
\*y_1(t_0, x) =  \sin(\pi x_1) \sin( \pi x_2), & \forall x\in \Omega,\\
\*y_2(t,x) = \begin{split}&t  \left(\sin(2 \pi x_1) \sin(2 \pi x_2) \right)^2\\& + (1-t) \sin(\pi x_1) \sin(\pi x_2)\end{split} & \forall  t \in [t_0,t_f], \forall x \in \Omega,\\
\*y_1(t, x) = \*y_2(t, x) = 0 & \forall x \in \partial \Omega,
\end{cases}
\end{split}
\end{equation}
where $\Omega = [0, 1]^2$, $t_0 = 0$, and $t_f = 1$.

\begin{figure}[H]
    \centering
    \includegraphics[width=0.7\textwidth]{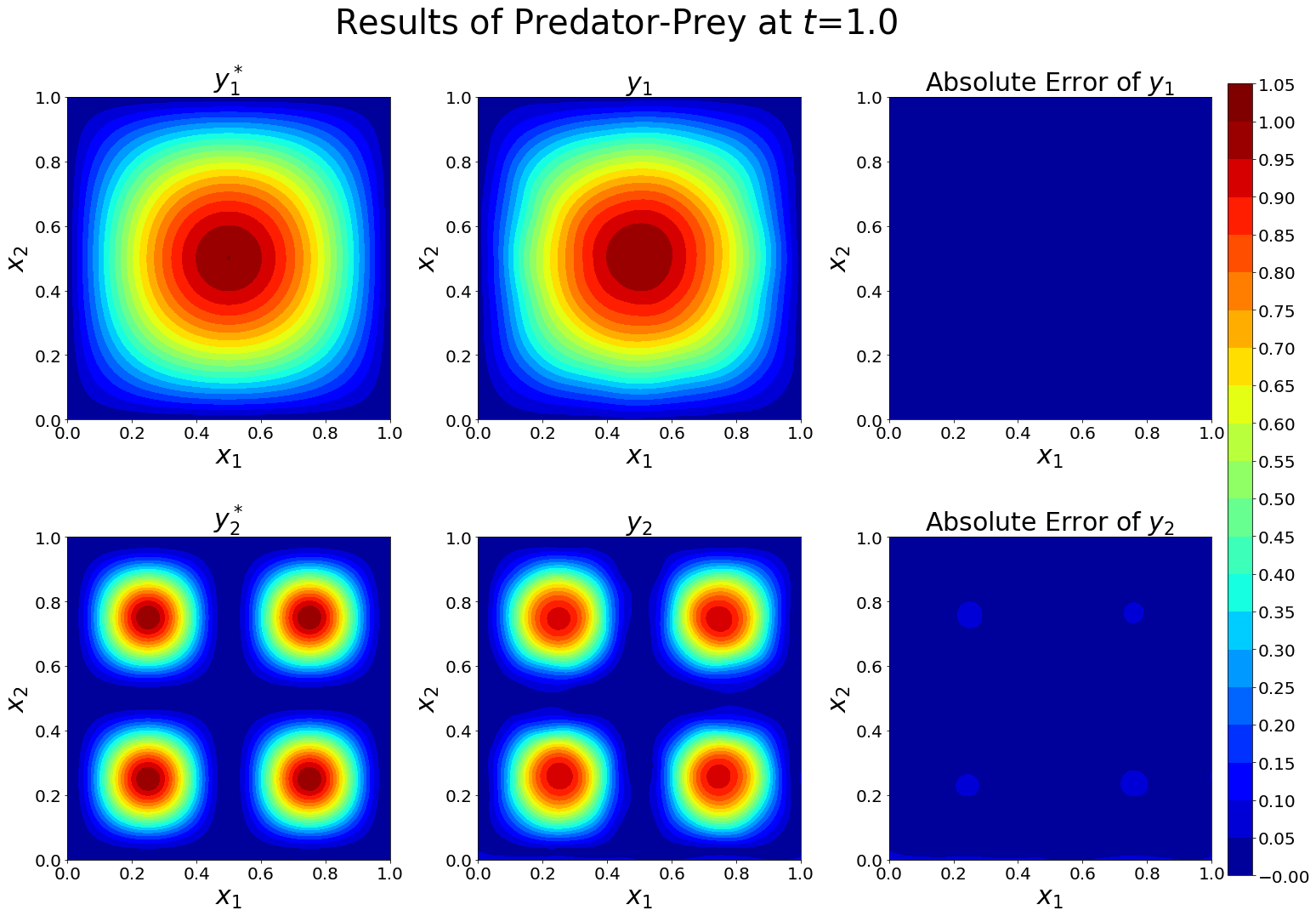}
    \caption{
    \textmd{\emph{Two-dimensional predator-prey results:} 
    Comparison of both the reference solution and the learned solution of the predator and prey populations at the final time $t=1.0$. 
    The reference solution of the predator population is denoted by $\*y_1^*$. 
    The learned solution of the predator population is denoted by $\*y_1$.
    The learned solution of the prey population is denoted by $\*y_2$. 
    The reference solution of the prey population is denoted by $\*y_2^*$.
    The right-most column displays the absolute error between the reference and learned solutions for each population.
    An expanded looked into the absolute errors throughout the timespan $t=[0, 1]$ is found in \cref{sec:Appendix-Predator-Prey}.}}
    \label{fig:predator-prey-absolute-error}
\end{figure}

\Cref{fig:predator-prey-absolute-error} offers a comparison of the reference solution of the both the prey and predator populations at the final time $t=1.0$. Additionally, \cref{fig:predator-prey-absolute-error} shows the absolute error between the reference and learned solutions for each population, illustrating how Control PINN has reached its objective at the end of the timespan. Please see \cref{sec:Appendix-Predator-Prey} for a more detailed look into the performance of Control PINN for this predator-prey problem. Moreover, an animation of the Control PINN model converging on the optimal solution can be found at 

Of note is the challenge faced by Control PINN to learn the boundary solution exactly, which is best characterized by the the three-dimensional modeling of this problem. In such a situation, the four concentrations within prey population are represented as cones. The base of these cones have a wavy boundary, indicating boundary enforcement issues. \cite{sifan2021deeponet} also makes note of their struggles to wrangle their framework's ability to balance solving the overall problem while maintaining satisfactory respect of the boundary conditions.

\biblio

%% file: sections/conclusion.tex
\section{Conclusions}
\label{sec:Conclusion}

This work provides a novel approach for solving the optimal  control problem for PDEs using PINNs. 
In contrast to previous approaches, we integrate the optimality conditions from the control problem directly in a theory-guided and physics-informed manner.
We dub this approach Control PINNs.

The Control PINN methodology is able to simultaneously learn the system state solution and the optimal control for a general class of PDEs.
We illustrate the validity of our approach on a diverse set of problems: a simple control problem for which an analytical solution is known, a one-dimensional heat equation for which a control that is not optimal is known, and a two-dimensional predator-prey problem which might not even have an optimal control. In higher dimensional problems, a tension exists in Control PINNs between respecting the boundary conditions of the given system state while learning the solution and corresponding optimal control. 
Adaptive methods for choosing the scaling of the terms in the loss similar to \cite[]{wang2021improvedarchitectures} are of future interest.

\textbf{Compute resources:} The GitHub repository for this paper is at 
\url{https://github.com/ComputationalScienceLaboratory/control-pinns}, and contains Google Colaboratory (Colab) notebooks for each experiment that are self-contained. 
The Colab notebook associated with the predator-prey problem in \cref{sec:2d-predator-prey} takes approximately forty-five minutes to run using a Tesla P100-PCIE-16GB GPU, uses 2.30 GB of RAM, and 38.74 GB of disk space. The other experiments are considerably faster to run, and less demanding resource-wise.

\textbf{Future work:} Future work will involve extending Control PINNs to solve closed-loop control problems. \cite{Nicodemus2021ModelPredictiveControl, Gokhale2022ControlOriented, LuLu2021HardConstraints} offer PINN approaches with which to compare and benchmark Control PINN's performance. Moreover, more complicated experiments such as Navier-Stokes will be utilized in subsequent complementary papers. 
Future results will be sure to offer a comparison between the adjoint and the PINN with no adjoint.
The computational costs, modes of failure, and scope of achievement will accompany said results.
After establishing Control PINN's viability in closed-loop control problems, one potential future application is leveraging Control PINNs as agents in deep reinforcement learning (DRL) \cite{cuomo2022scientific}. In DRL, finding the state of a robot after a given action requires solving a number of physical equations (e.g. equation of motion and balance of force). This issue can be circumvented by leveraging PINNs as an agent because PINNs penalize deviations from physical constraints by design. Furthermore, an agent that can simultaneously solve a system state and a corresponding optimal control, such as Control PINN, would be useful in Q-learning to efficiently optimize the value of action-state policies. 

\textbf{Broader impacts:} A more efficient way to model and optimally control machines interacting with the real world has various societal impacts. Faster modeling begets faster understanding, and consequently a positive feedback loop can be established regarding answering certain questions of interest. With regard to potential negative impacts on society, offloading multistage decision processes to machines in the form of an optimal control policy lends credence to the idea of a future reduction in the demand of human labor. Energy resource optimization, industrial automation, and urban complex systems could be affected from combining optimal control and machine learning in the form of Control PINNs.

\biblio

%% file: sections/appendix_analytical_problem.tex
\section{Analytical problem}
\label{sec:Appendix-Analytical-Problem}

\begin{figure}[H]
    \centering
    \includegraphics[height=0.85\textheight]{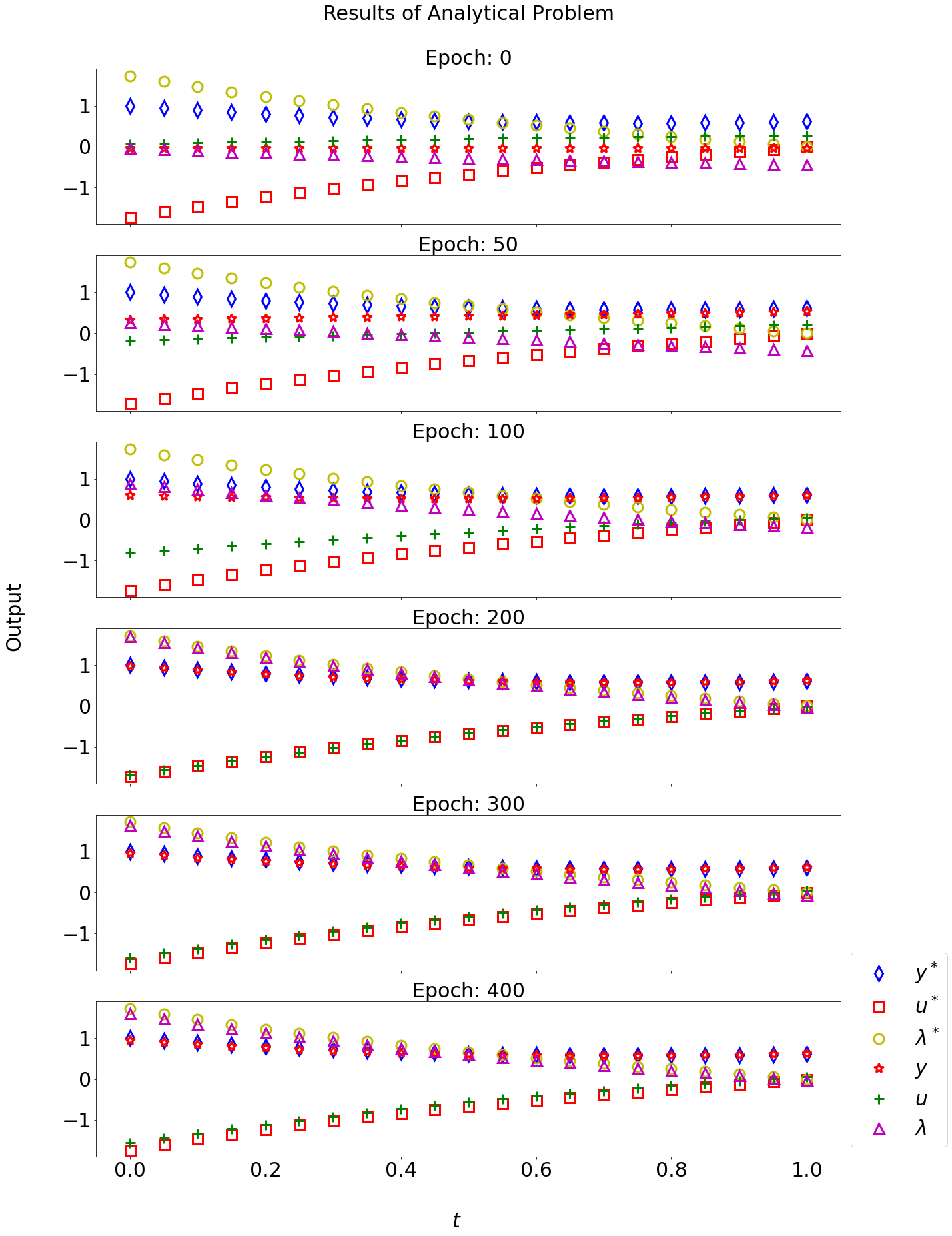}
    \caption{\textmd{\emph{Results of the analytical problem:} The optimal solutions of the system state, control, and adjoint on the control are respectively denoted by $\*y^*$, $\*u^*$, and $\*\lambda^*$. Their corresponding learned solutions found by the Control PINN model are denoted by $\*y$, $\*u$, and $\*\lambda$. After 300 epochs Control PINN has reached convergence on the analytical solution, and the solutions remain unchanged upon further training.}}
    \label{fig:1d-heat-expanded-results}
\end{figure}

\begin{figure}[H]
    \centering
    \includegraphics[]{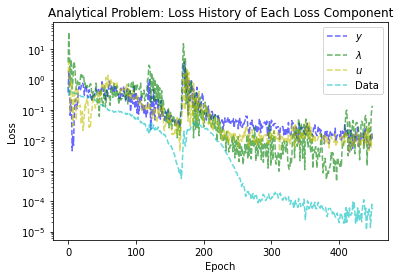}
    \caption{\textmd{\emph{History for each loss component of the analytical problem.}}}
    \label{fig:analytical-problem-loss-history}
\end{figure}

%% file: sections/appendix_heat_equation.tex
\section{One-dimensional heat equation}
\label{sec:Appendix-Heat-Equation}

\begin{figure}[H]
    \centering
    \includegraphics[width=0.85\textwidth]{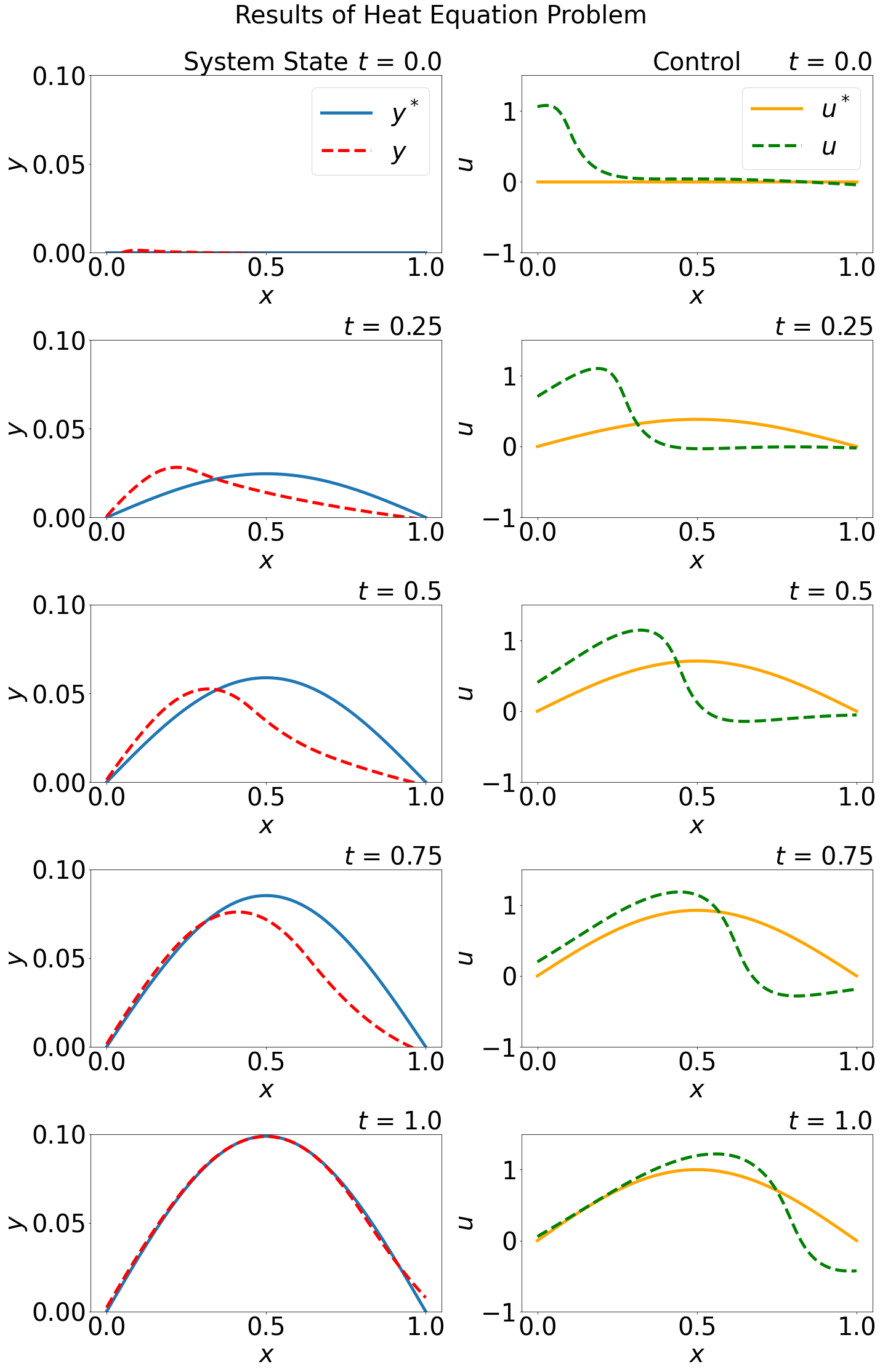}
    \caption{\textmd{\emph{One-dimensional heat equation results:} \emph{Left:} Comparison of the learned system state $\*y$ and the reference system state $\*y^*$. \emph{Right:} Comparison of the learned control $\*u$ and the reference control $\*u^* = sin(\pi x) sin(\frac{\pi t}{2})$. Note that there are multiple control signals that satisfy the PDE solution. The reference control $\*u^*$ might not be the optimal solution, and is not the same as the control signal found by Control PINN. Control PINN found a different control that minimizes the control action on the system state.}}
    \label{fig:1d-heat-expanded-results}
\end{figure}

\begin{figure}[H]
    \centering
    \includegraphics[width=0.85\textwidth]{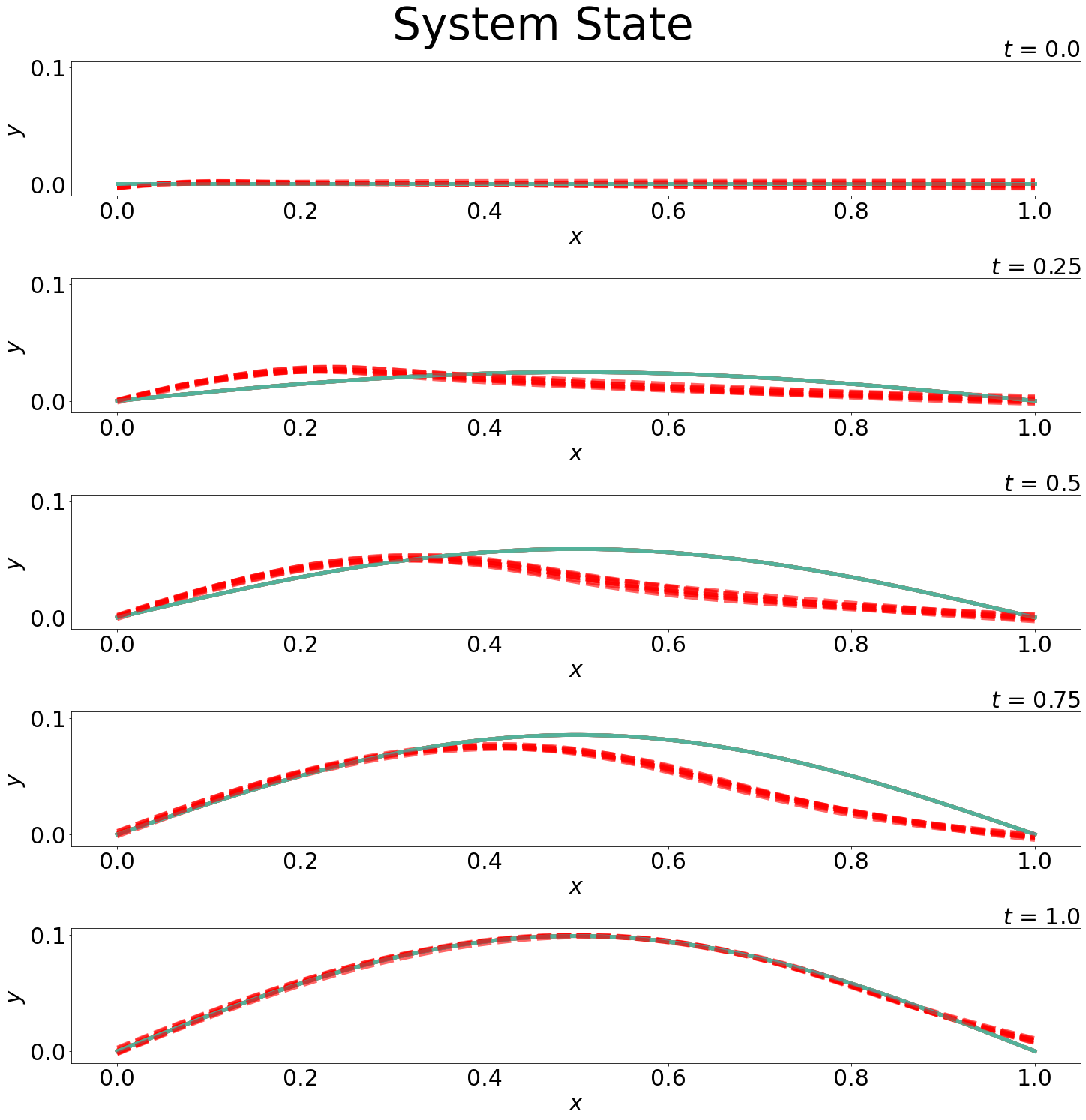}
    \caption{\textmd{\emph{One-dimensional heat equation solution convergence on system state:} The heat equation was trained on ten different random initialization of the weights, which resulted in learning the same approximate system state solution.}}
    \label{fig:1d-heat-system-random-seed}
\end{figure}

\begin{figure}[H]
    \centering
    \includegraphics[width=0.85\textwidth]{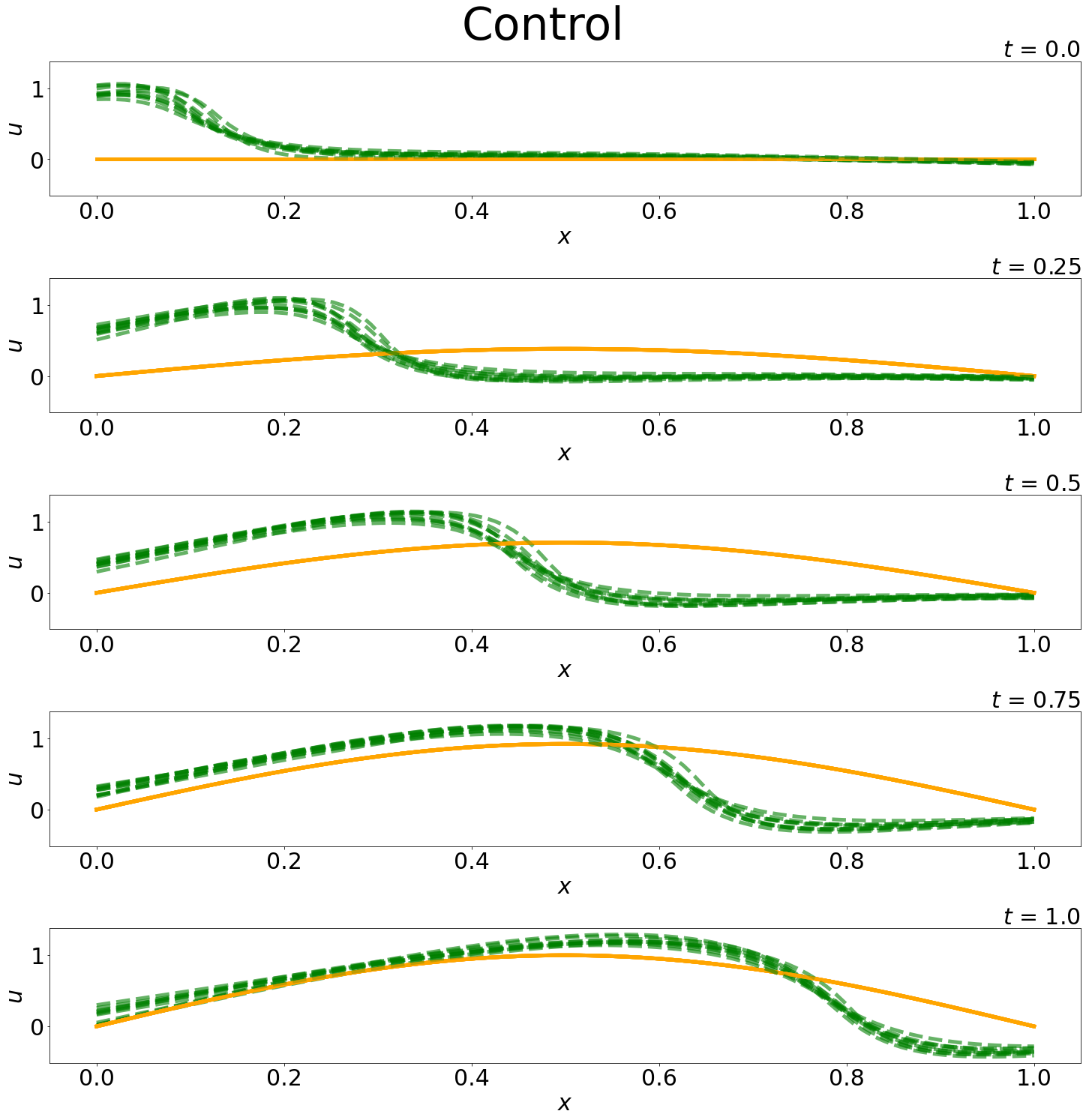}
    \caption{\textmd{\emph{One-dimensional heat equation solution convergence on control signal:} The heat equation was trained on ten different random initialization of the weights, which resulted in learning the same approximate control signal solution.}}
    \label{fig:1d-heat-control-random-seed}
\end{figure}

\begin{figure}[H]
    \centering
    \includegraphics{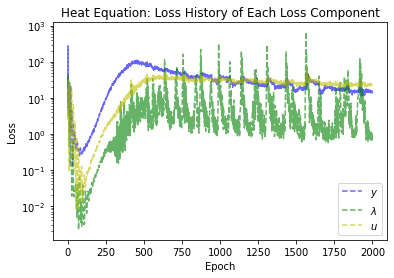}
    \caption{\textmd{\emph{History for each loss component of the one-dimensional heat equation.} }}
    \label{fig:1d-heat-loss-history-components}
\end{figure}

\begin{figure}[H]
    \centering
    \includegraphics{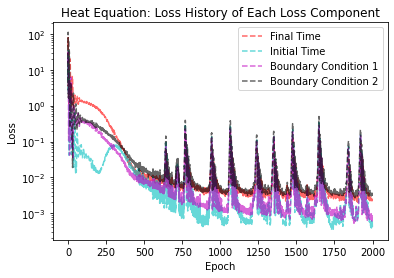}
    \caption{\textmd{\emph{History for the boundary conditions of the one-dimensional heat equation.} }}
    \label{fig:1d-heat-loss-history-bc}
\end{figure}

%% file: sections/appendix_predator_prey.tex
\section{Two-dimensional predator-prey problem}
\label{sec:Appendix-Predator-Prey}

\begin{figure}[H]
    \centering
    \includegraphics[width=\textwidth]{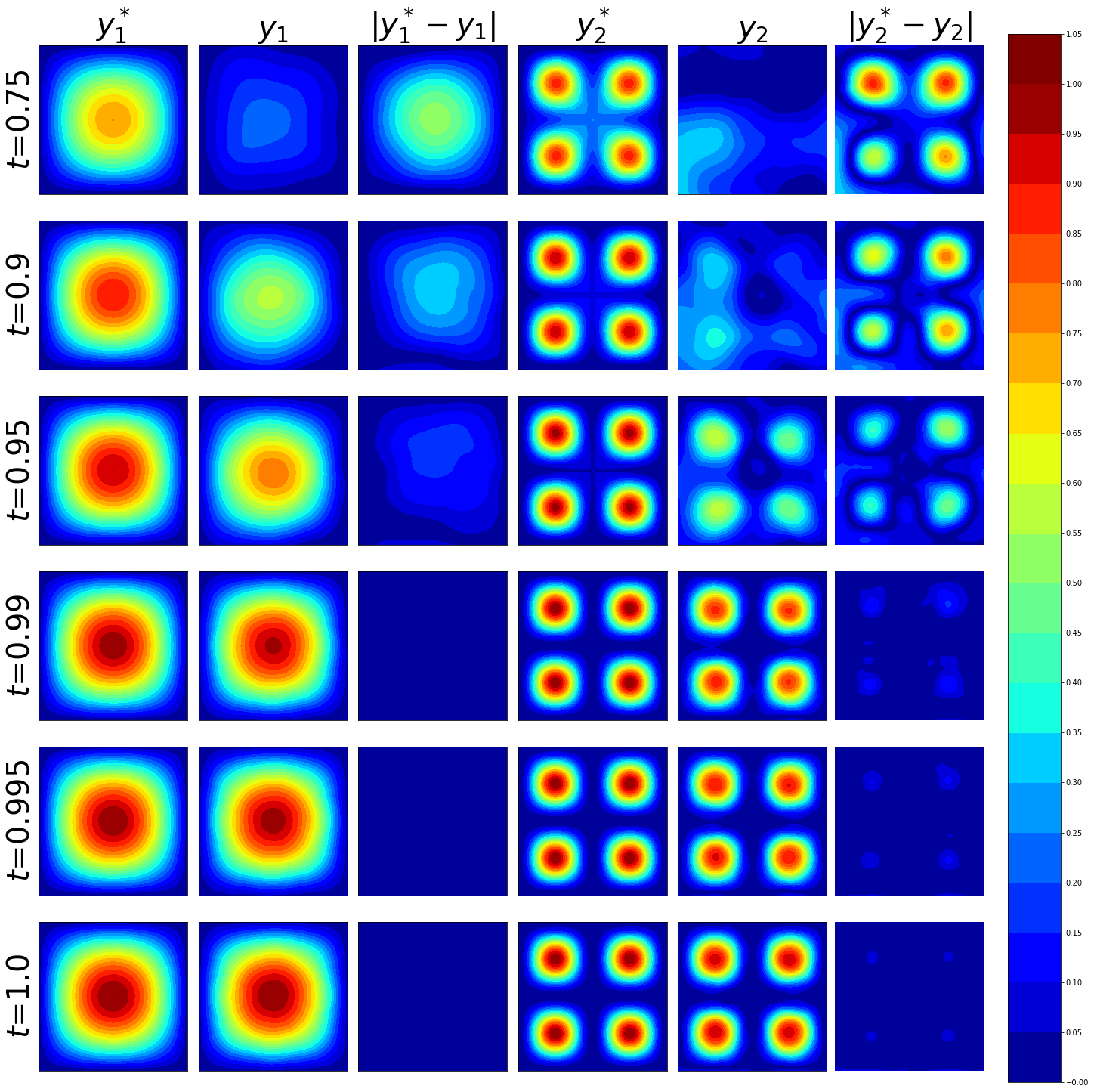}
    \caption{\textmd{\emph{Two-dimensional predator-prey results:} 
    Expanded comparison of the absolute error between the reference and learned solutions for the timepoints $t \in [0.75, 0.9, 0.95, 0.99, 0.995, 1.0]$, with each row in the plot denoting a given point in the timespan. 
    \emph{First column:} The reference solution $\*y_1^*$ of the system, and its corresponding state respective to the given timepoint row.
    \emph{Second column:} The learned solution $\*y_1$ of the system by Control PINN, and its corresponding state respective to the given timepoint row.
    \emph{Third column:} The absolute error between the reference and learned solutions of the predator populations ($\*y_1^*$ and $\*y_1$, respectively). 
    \emph{Fourth column:} The reference solution $\*y_2^*$ of the system, and its corresponding state respective to the given timepoint row.
    \emph{Fifth column:} The learned solution $\*y_2$ of the system by Control PINN, and its corresponding state respective to the given timepoint row.
    \emph{Sixth column:} The absolute error between the reference and learned solutions of the prey populations ($\*y_2^*$ and $\*y_2$, respectively).
    \vspace{8em}
    }}
    \label{fig:predator-prey-absolute-error-apx}
\end{figure}

\begin{figure}[H]
    \centering
    \includegraphics[width=0.7\linewidth]{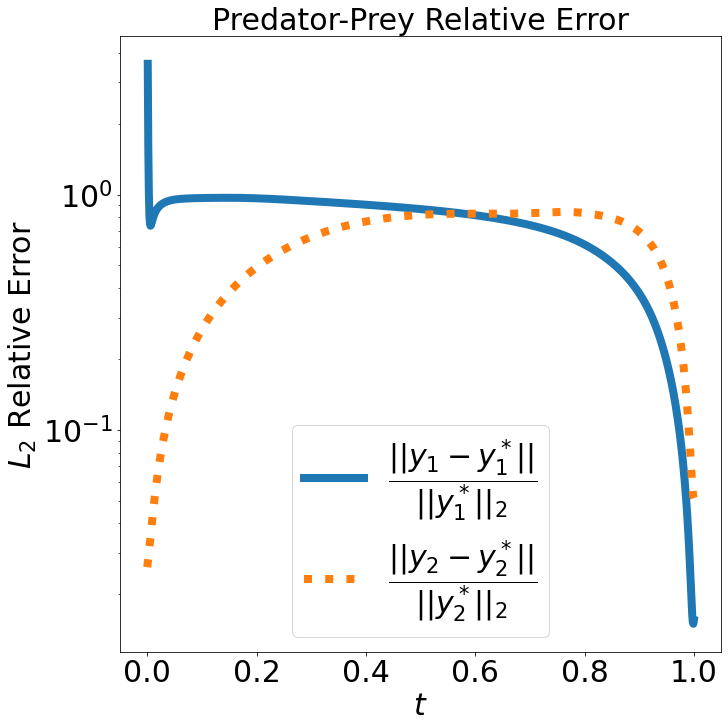}
    \caption{\textmd{\emph{Two-dimensional predator-prey problem:} 
    The $L_2$ norm relative error of both the learned predator and prey populations. The reference solution of the predator population is denoted by $\*y_1^*$. The reference solution of the prey population is denoted by $\*y_2^*$. The learned solution of the predator population by Control PINN is denoted by $\*y_1$. The learned solution of the prey population is by Control PINN is denoted by $\*y_1$.
    }}
    \label{fig:predator-prey-l2-relative-error}
\end{figure}

\begin{figure}[H]
    \centering
    \includegraphics[width=\linewidth]{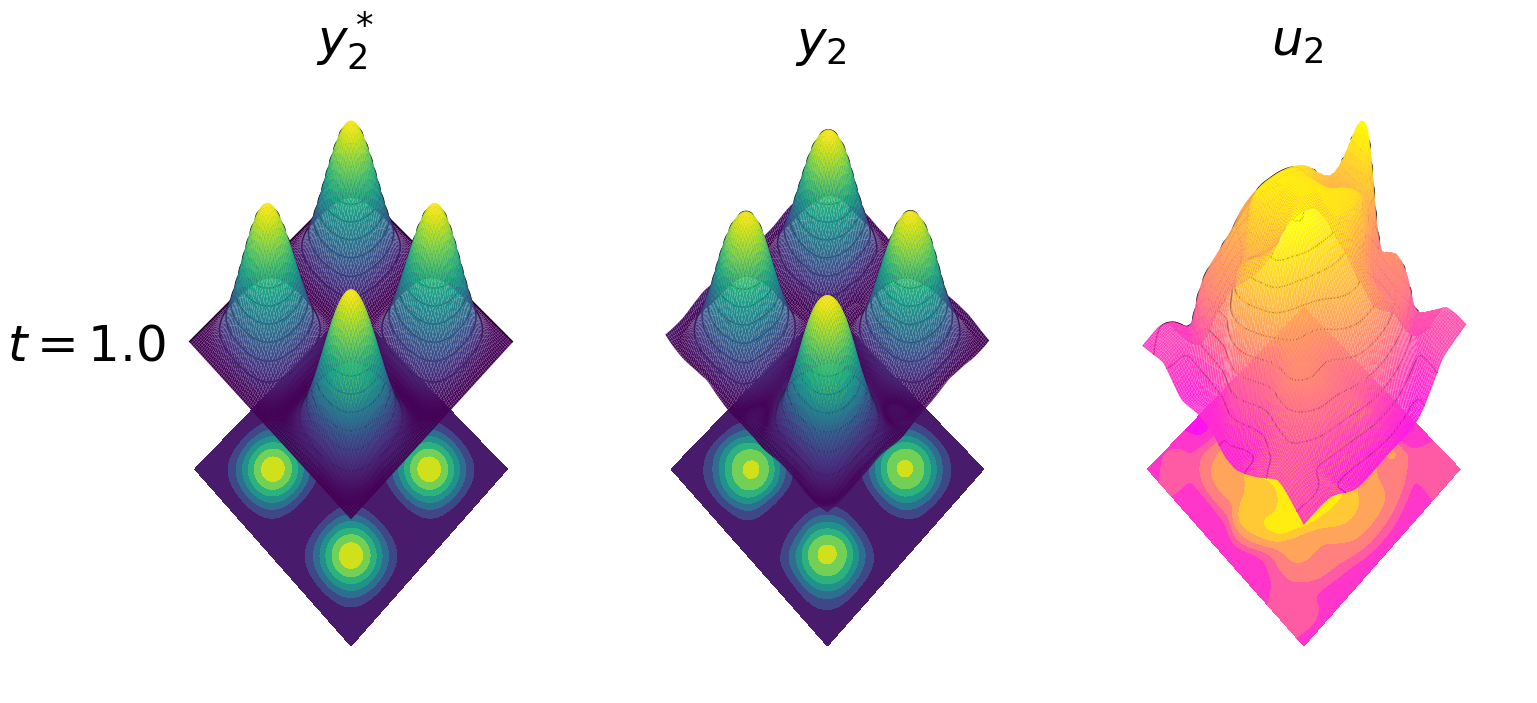}
    \caption{\textmd{\emph{Two-dimensional predator-prey problem:} \emph{Left:} Prescribed prey population at the final time $t_f=1.0$. \emph{Middle:} Learned solution $\*y_2$ prey population at the final time $t_f=1.0$. \emph{Right:} Control signal $\*u_2$ on the predator population $\*y_2$ determined by the Control PINN at the final time $t_f=1.0$. The three-dimensional surface plot is projected downward to a two-dimensional contour plot for increased discrepancy of comparative purposes. Note the differences in the bases of the surface plots, which illustrates the challenges of boundary condition enforcement.}}
    \label{fig:ControlPINN_FinalTime_PredatorComparison}
\end{figure}

\begin{figure}[H]
    \centering
    \includegraphics[width=\linewidth]{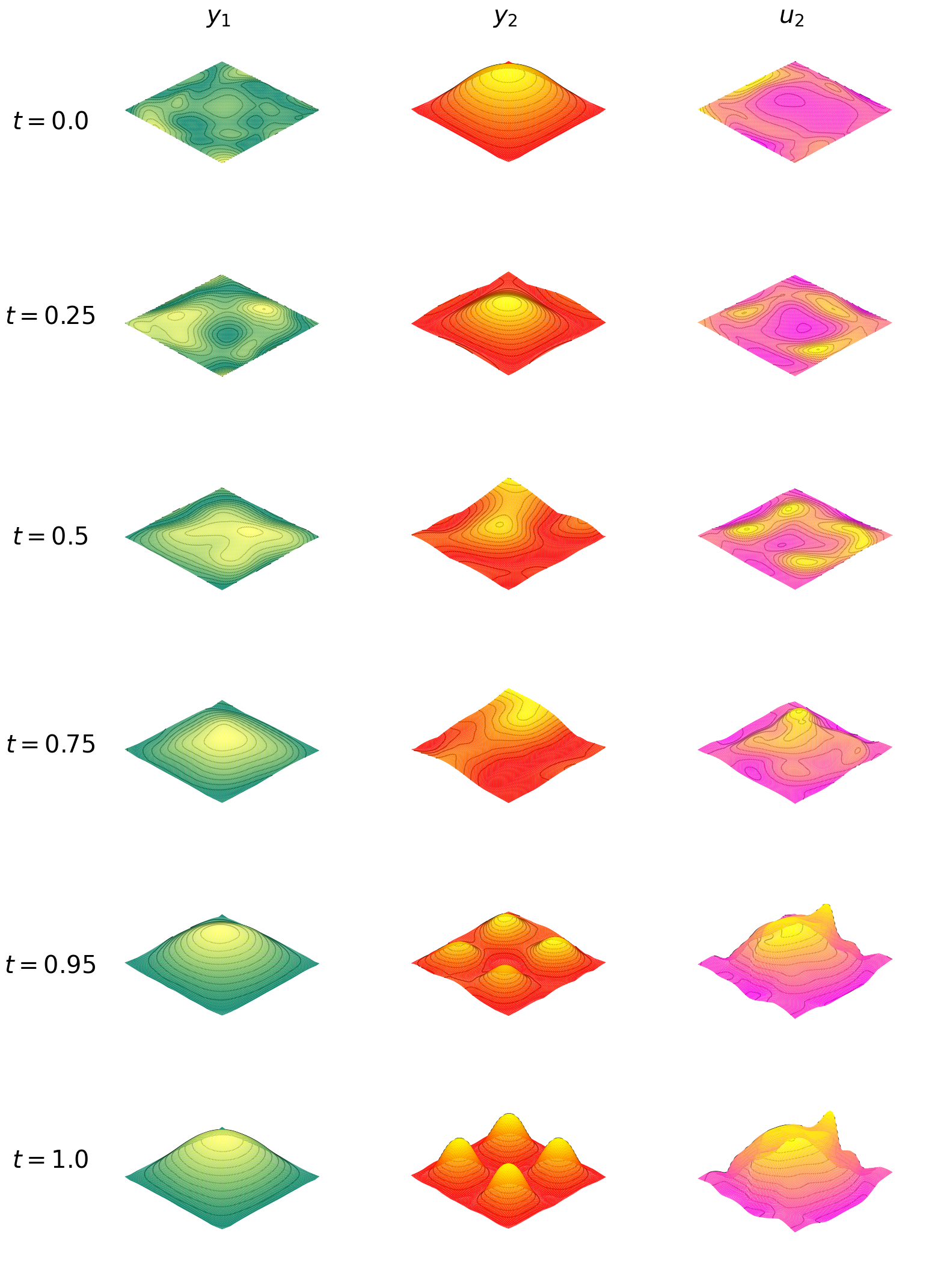}
    \caption{\textmd{\emph{Two-dimensional predator prey problem:} Surface plots of the learned solutions for the predator population, prey population, and control at varying time points.
    \emph{First column:} The learned solution of the predator population by Control PINN, denoted by $\*y_1$.
    \emph{Second column:} The learned solution of the prey population by Control PINN, denoted by $\*y_2$.
    \emph{Third column:} The learned solution of the control signal by Control PINN, denoted by $\*u_2$.
    Control PINN minimizes the control needed to reach the desired population states by waiting until near the end to enact significant control on the prey population.
    }}
    \label{fig:predator-prey-3d-surface-expanded-results}
\end{figure}

\begin{figure}[H]
    \centering
    \includegraphics{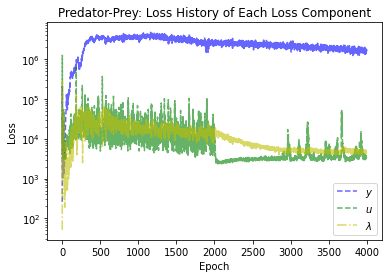}
    \caption{\textmd{\emph{History for each loss component of the two-dimensional predator-prey problem.} }}
    \label{fig:2d-predator-prey-history-components}
\end{figure}

\begin{figure}[H]
    \centering
    \includegraphics{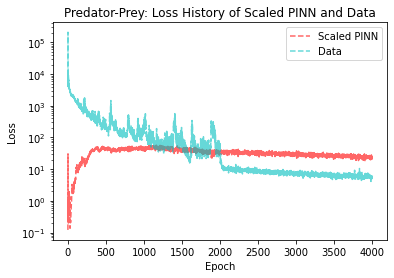}
    \caption{\textmd{\emph{History for the scaled PINN and data loss of the two-dimensional predator-prey problem.} }}
    \label{fig:2d-predator-prey-history-bc}
\end{figure}